\documentclass[sigconf]{acmart}

\usepackage{multirow}
\usepackage{makecell}
\usepackage{booktabs}
\usepackage{subfigure}
\usepackage{enumitem}
\usepackage{fancyhdr}

\def\BibTeX{{\rm B\kern-.05em{\sc i\kern-.025em b}\kern-.08emT\kern-.1667em\lower.7ex\hbox{E}\kern-.125emX}}

\setcopyright{acmcopyright}

\copyrightyear{2019} 
\acmYear{2019} 
\acmConference[CIKM '19]{The 28th ACM International Conference on Information and Knowledge Management}{November 3--7, 2019}{Beijing, China} 
\acmBooktitle{The 28th ACM International Conference on Information and Knowledge Management (CIKM '19), November 3--7, 2019, Beijing, China} 
\acmPrice{15.00} 
\acmDOI{10.1145/3357384.3357943} 
\acmISBN{978-1-4503-6976-3/19/11}

\begin{document}
%
%
\title{Temporal Network Embedding with Micro- and Macro-dynamics}

%

%
%

\author{Yuanfu Lu}
\affiliation{%
  \institution{Beijing University of Posts and Communications}
  \city{Beijing}
  \country{China}}
\email{luyuanfu@bupt.edu.cn}

\author{Xiao Wang}
\affiliation{%
  \institution{Beijing University of Posts and Communications}
  \city{Beijing}
  \country{China}}
\email{xiaowang@bupt.edu.cn}
 
\author{Chuan Shi}
\authornote{Corresponding author.}
\affiliation{%
  \institution{Beijing University of Posts and Communications}
  \city{Beijing}
  \country{China}}
\email{shichuan@bupt.edu.cn}
 
\author{Philip S. Yu}
\affiliation{%
  \institution{University of Illinois at Chicago}
  \city{IL}
  \country{USA}}
\email{psyu@uic.edu}
 
\author{Yanfang Ye}
\affiliation{%
  \institution{Case Western Reserve University}
  \city{OH}
  \country{USA}}
\email{yanfang.ye@case.edu}


%
\begin{abstract}
Network embedding aims to embed nodes into a low-dimensional space, while capturing the network structures and properties. Although quite a few promising network embedding methods have been proposed, most of them focus on static networks.  
In fact, temporal networks, which usually evolve over time in terms of microscopic and macroscopic dynamics, are ubiquitous. The micro-dynamics describe the formation process of network structures in a detailed manner, while the macro-dynamics refer to the evolution pattern of the network scale. 
Both micro- and macro-dynamics are the key factors to network evolution; however, how to elegantly capture both of them for temporal network embedding, especially macro-dynamics, has not yet been well studied. 
In this paper, we propose a novel temporal network embedding method with micro- and macro-dynamics, named $\rm{M^2DNE}$. 
Specifically, for micro-dynamics, we regard the establishments of edges as the occurrences of chronological events and propose a temporal attention point process to capture the formation process of network structures in a fine-grained manner. 
For macro-dynamics, we define a general dynamics equation parameterized with network embeddings to capture the inherent evolution pattern and impose constraints in a higher structural level on network embeddings. 
Mutual evolutions of micro- and macro-dynamics in a temporal network alternately affect the process of learning node embeddings. 
Extensive experiments on three real-world temporal networks demonstrate that $\rm{M^2DNE}$ significantly outperforms the state-of-the-arts not only in traditional tasks, e.g., network reconstruction, but also in temporal tendency-related tasks, e.g., scale prediction. 
\end{abstract}
\keywords{Network Embedding, Temporal Network, Social Dynamics Analysis, Temporal Point Process}

\maketitle

\begin{figure}
	\includegraphics[width=\linewidth]{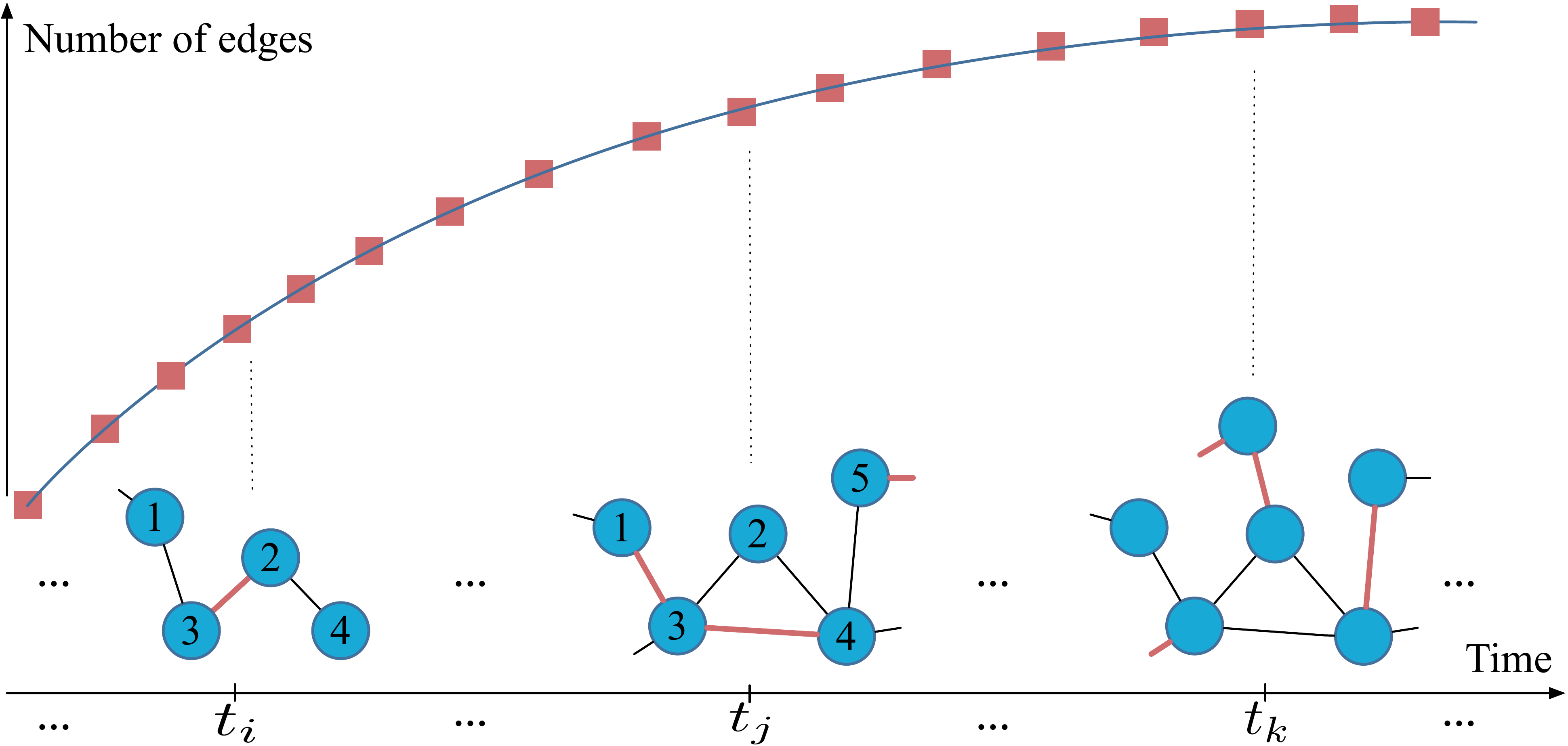}
	\caption{A toy example of temporal network with micro- and macro-dynamics. Micro-dynamics describe the formation process of network structures (i.e., edge establishments), while macro-dynamics refer to the evolution pattern of network scale (i.e., the number of edge). The red lines in the network mean new temporal edges (e.g., $(v_3,v_4,t_j)$).}
	\label{fig:problem_fig}
\end{figure}
\section{Introduction}
Network embedding has shed a light on network analysis due to its capability of encoding the structures and properties of networks with latent representations \cite{cui2018survey,cai2018comprehensive}. Though the state-of-the-arts \cite{perozzi2014deepwalk, tang2015line,wang2016structural,grover2016node2vec,dong2017metapath2vec,qiu2018network} have achieved promising performance in many data mining tasks, most of them focus on static networks with fixed structures.  
In reality, the network usually exhibits complex temporal properties, meaning that the network structures are not achieved overnight and usually evolve over time. In the so-called temporal networks \cite{Li1042}, the establishments of edges between nodes are chronological and the network scale grows with some obvious distribution. For example, researchers collaborate with others in different years, leading to sequential co-author events and continued growth of the network scale.  
Therefore, a temporal network naturally represents the evolution of a network, including not only the fine-grained network structure but also the macroscopic network scale. 
Embedding a temporal network with the latent representation space is of great importance for applications in practice. 

Basically, one requirement for temporal network embedding is that the learned embeddings should preserve the network structure and reflect its temporal evolution. 
The temporal network evolution usually follows two dynamics processes, i.e., the microscopic and macroscopic dynamics. 
At the microscopic level, the temporal network structure is driven by the establishments of edges, which is actually a sequence of chronological events involving two nodes. Taking Figure \ref{fig:problem_fig} as an example, from time $t_i$ to $t_j$, the formation of the structure can be described as $\{(v_3,v_1,t_i)$, $(v_3,v_2,t_i)$, $\cdots$$\}$ $\xrightarrow{...}$$\{(v_3,v_1,t_j),(v_3,v_4,t_j),\cdots\}$, and here nodes $v_3$ and $v_1$ build a link again at $t_j$. 
Usually, an edge generated at time $t$ is inevitably related with the historical neighbors before $t$, and the influence of the neighbor structures on the edge formation comes from the two-way nodes, not just a single one. Besides, different neighborhoods may have distinct influences. 
 For example, in Figure \ref{fig:problem_fig}, the establishment of edge $(v_3,v_4)$ at time $t_j$ should be influenced by \{$(v_3,v_1,t_i)$, $(v_3,v_2,t_i),\cdots\}$ and $\{(v_4,v_2,t_i),\cdots\}$. Besides, the influences of $(v_3,v_2,t_i)$ and $(v_4,v_2,t_i)$ on the event $(v_3,v_4,t_j)$ should be larger than that of $(v_3,v_1,t_i)$, since nodes $v_2, v_3$ and $v_4$ form a closed triad \cite{zhou2018dynamic,HuangTLLF15}. 
Such micro-dynamic evolution process describes the edge formation between nodes at different timesteps in detail and explains that ``why the network evolves into such structures at time $t$''. Modeling the micro-dynamics enables the learned node embeddings to capture the evolution of a temporal network more accurately, which will be beneficial for the downstream temporal network tasks. 
We notice that temporal network embedding has been studied by some works \cite{du2018dynamic, zhou2018dynamic, trivedi2018representation, zuo2018embedding}. However, they either simplify the evolution process as a series of network snapshots, which cannot truly reveal the formation order of edges; or model neighborhood structures using stochastic processes, which ignores the fine-grained structural and temporal properties. 

More importantly, at the macroscopic level, another salient property of the temporal network
is that the network scale evolves with obvious distributions over time, e.g., S-shaped sigmoid curve \cite{leskovec2005graphs} or a power-law like pattern \cite{zang2016beyond}. 
As shown in Figure \ref{fig:problem_fig}, when the network evolves over time, the edges are continuously being built and form the network structures at each timestamp. 
Thus, the network scale, i.e., the number of edges, grows with time and obeys a certain underlying principle, rather than being randomly generated. 
Such macro-dynamics reveal the inherent evolution pattern of the temporal network and impose constraints in a higher structural level on the network embedding, i.e., they determine that how many edges should be generated totally by micro-dynamics embedding as the network evolves. 
The incorporation of macro-dynamics provides valuable and effective evolutionary information to enhance the capability of network embedding preserving network structure and evolution pattern, which will largely strengthen the generalization ability of network embedding. 
Therefore, whether the learned embedding space can encode the macro-dynamics in a temporal network should be a critical requirement for temporal network embedding methods. 
Unfortunately, none of the existing temporal network embedding method takes them into account although the macro-dynamics are closely related with temporal networks.  

In this paper, we propose a novel temporal \textbf{N}etwork \textbf{E}mbedding method with \textbf{M}icro- and \textbf{M}acro-\textbf{D}ynamics, named $\mathbf{M^2DNE}$.   
In particular, to model the chronological events of edge establishments in a temporal network (i.e., micro-dynamics), we elaborately design a \textit{temporal attention point process} by parameterizing the conditional intensity function with node embeddings, which captures the fine-grained structural and temporal properties with a \textit{hierarchical temporal attention}. 
To model the evolution pattern of the temporal network scale (i.e., macro-dynamics), we define a general \textit{dynamics equation} as a non-linear function of the network embedding, which imposes constraints on the network embedding at a high structural level and well couples the dynamics analysis with representation learning on temporal networks. 
At last, we combine micro- and macro-dynamics preserved embedding and optimize them jointly. 
As micro- and macro-dynamics mutually evolve and alternately influence the process of learning node embeddings, the proposed $\rm{M^2DNE}$ has the capability to capture the formation process of topological structures and the evolutionary pattern of network scale in a unified manner. We will make our code and data publicly available at website after the review.

The major contributions of this work can be summarized as follows:
\begin{itemize}[leftmargin=*]
	\item For the first time, we study the important problem of incorporating the micro-dynamics and macro-dynamics into temporal network embedding.
	\item We propose a novel temporal network embedding method ($\rm{M^2DNE}$), which microscopically models the formation process of network structure with a temporal attention point process, and macroscopically constrains the network structure to obey a certain evolutionary pattern with a dynamics equation.
	\item We conduct comprehensive experiments to validate the benefits of $\rm{M^2DNE}$ on the traditional applications (e.g., network reconstruction and temporal link prediction), as well as some novel applications related to temporal networks (e.g. scale prediction).
\end{itemize}

\section{Related work}
Recently, network embedding has attracted considerable attention \cite{cui2018survey}. Inspired by word2vec \cite{mikolov2013distributed}, random walk based methods\cite{perozzi2014deepwalk,grover2016node2vec} have been proposed to learn node embeddings by the skip-gram model. 
After that, \cite{wang2017community,qiu2018network} are designed to better preserve network properties, e.g. high-order proximity. 
There are also some deep neural network based methods, such as autoencoder based methods \cite{wang2016structural,wang2018shine} and graph neural network based methods \cite{KipfW17,gat}.  
Besides, some models are designed for heterogeneous information networks \cite{dong2017metapath2vec,shi2018heterogeneous,LuSH019} or attribute networks \cite{ijcai2018-438}. However, all the aforementioned methods only focus on  static network embedding. 

There are some attempts in temporal network embedding, which can be broadly classified into two categories: embedding snapshot networks \cite{li2017attributed, zhu2018high, du2018dynamic,goyal2018dyngem, zhou2018dynamic} and modeling temporal evolution \cite{nguyen2018continuous,trivedi2018representation,zuo2018embedding}. 
The basic idea of the former is to learn node embedding for each network snapshot.  Specifically, DANE \cite{li2017attributed} and DHPE \cite{zhu2018high} present efficient algorithms based on perturbation theory. Song et al. extend skip-gram based models and propose a dynamic network embedding framework \cite{du2018dynamic}. DynamicTriad \cite {zhou2018dynamic} models the triadic closure process to capture dynamics and learns node embeddings at each time step. 
The latter type of methods try to capture the evolution pattern of network for latent embeddings. \cite{trivedi2018representation} describes temporal evolution over graphs as association and communication process and propose a deep representation learning framework for dynamic graphs. HTNE \cite{zuo2018embedding} proposes a Hawkes process based network embedding method, which models the neighborhood formation sequence to learn node embeddings. 
Besides, there are some task-specific temporal network embedding methods. 
NetWalk \cite{yu2018netwalk} is an anomaly detection framework, which detects network deviations based on a dynamic clustering algorithm. 

All the above-mentioned methods either learn node embeddings on snapshots, or model temporal process of networks with limited dynamics and structures. 
None of them integrate both of micro- and macro-dynamics into temporal network embedding. 

\begin{figure*}
	\includegraphics[width=0.92\linewidth]{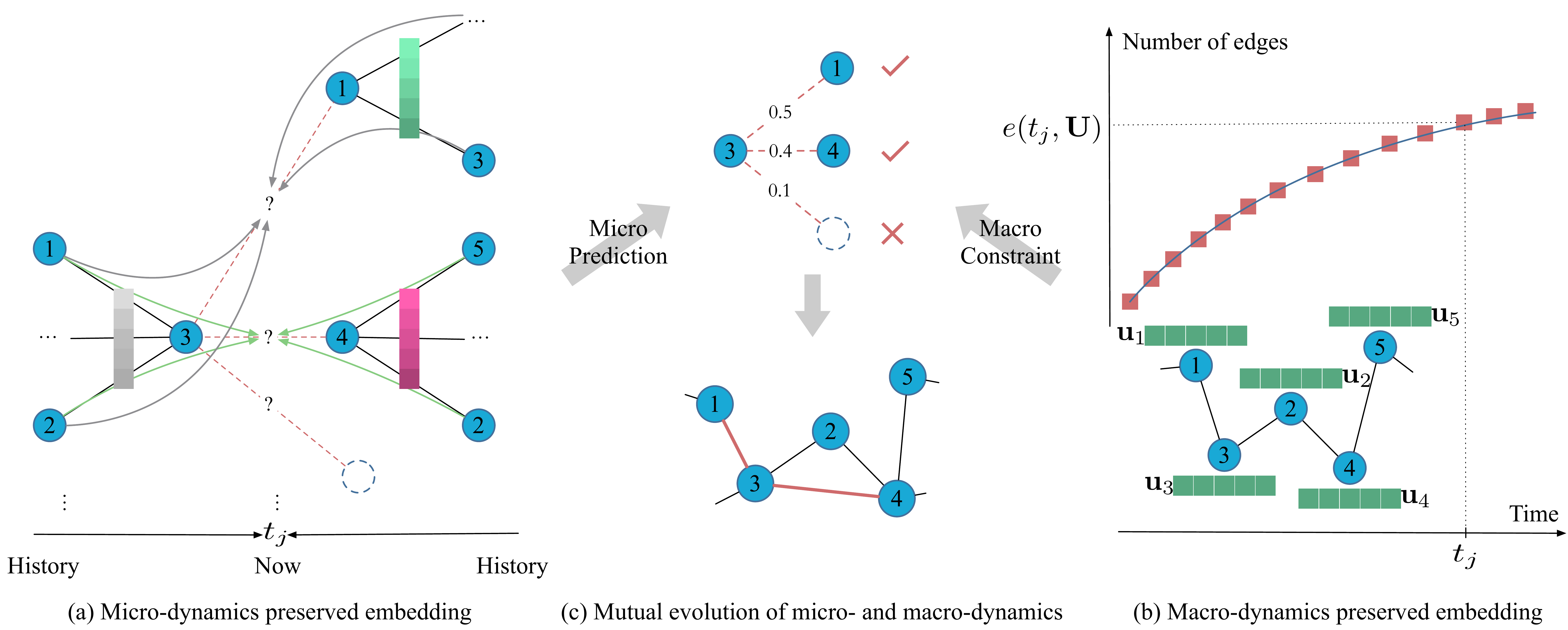}
	\Description{model}
	\caption{The overall architecture of $\mathbf{M^2DNE}$. 
	(a) Micro-dynamics preserved embedding with a temporal attention point process. The dash lines indicate the edges to be established and the solid arrows with different colors indicate the influence of neighbors on different edges. Vertical colored rectangular blocks represent attention mechanism, and the darker the color, the larger the influence of neighbors. 
	(b) Macro-dynamics preserved embedding with a dynamics equation, which is parameterized with network embedding $\mathbf{U}$ (i.e., green rectangular blocks) and time $t$. 
	At time $t_j$, $\mathbf{M^2DNE}$ macroscopically constraints the number of edges to $e(t_j,\mathbf{U})$. 
	(c) Micro- and macro-dynamics evolve and derive node embeddings in a mutual manner. 
	At the current time $t_j$, by microscopically predicting from historical neighbors (i.e., $\{v_1,v_2,\cdots\}$ and $\{v_5,v_2,\cdots\}$), $v_3$ links nodes $v_1$, $v_4$ and $v_?$ with the probability of $0.5$, $0.4$, and $0.1$, respectively; while macro-dynamics limit the number of new edges to only 2, according to the evolution pattern of the network scale. 
	 As a result, $\mathbf{M^2DNE}$ captures more precise structural and temporal properties.} 
	\label{fig:model_fig}
\end{figure*}
\section{Preliminaries}
\subsection{Dynamics in Temporal Networks}
\begin{definition}\textbf{Temporal Network}. 
	A temporal network refers as a sequence of timestamped edges, where each edge connects two nodes at a certain time. Formally, a temporal network can be denoted as $\mathcal{G}=(\mathcal{V}, \mathcal{E}, \mathcal{T})$, where $\mathcal{V}$ and $\mathcal{E}$ are sets of nodes and edges, and $\mathcal{T}$ is the sequence of timestamps. Each temporal edge $(i, j, t) \in \mathcal{E}$ refers to an event involving nodes $i$ and $j$ at time $t$. 
\end{definition}

Please notice that nodes $i$ and $j$ may build multiple edges at different timestamps, we consider $(i,j,t)$ as a distinct temporal edge while $(i,j)$ means a static edge. 
\begin{definition}\textbf{Micro-dynamics}.
	Given a temporal network $\mathcal{G}=(\mathcal{V}, \mathcal{E}, \mathcal{T})$, micro-dynamics describe the formation process of the network structures, denoted as $\mathcal{I} = \{(i, j, t)_m\}_{m=1}^{|\mathcal{E}|}$, where $(i, j, t)$ represents a temporal event that nodes $i$ and $j$ establish an edge at time $t$ and $\mathcal{I}$ is the complete sequence of $|\mathcal{E}|$ observed events ordered by time in window $[0, \mathcal{T}]$.
\end{definition}

\begin{definition}\textbf{Macro-dynamics}. 
	Given a temporal network $\mathcal{G}=(\mathcal{V}, \mathcal{E}, \mathcal{T})$, macro-dynamics refer to the evolution process of the network scale, denoted as $\mathcal{A}= \{e(t)\}_{t=t_1}^{|\mathcal{T}|}$, where $e(t)$ is the cumulative number of edges by time $t$. 
\end{definition}

In fact, macro-dynamics represent both the change of edges and nodes. Since new nodes will inevitably lead to new edges, we focus on the growth of edges here.  
Intuitively, micro-dynamics determine which edges will be built (i.e., events occurrence), while macro-dynamics constrain the scale of new changes of edges.

\subsection{Temporal Point Process}
Temporal point processes have previously been used to model dynamics in networks \cite{NIPS2015_5754}, which assumes that an event happens in a tiny window $[t, t + dt)$ with conditional probability $\lambda(t)dt$ given the historical events. Self-exciting  multivariate point process or Hawkes process \cite{mei2017neural} is a well-known temporal point process with the conditional intensity function $\lambda(t)$ defined as follows: 
\begin{equation}
\begin{small}
	\lambda(t) = \mu(t) + \int_{-\infty}^t\kappa(t-s)dN(s),
\end{small}
\end{equation} 
where $\mu(t)$ is the base intensity, describing the arrival of spontaneous events. The kernel $\kappa (t-s)$ models the time decay effects of past events on current events, which is usually in the form of an exponential function. $N(t)$ is the number of events until $t$.

\subsection{Problem Definition}
Our goal is to learn node embeddings by capturing the formation process of the network structures and the evolution pattern of the network scale. We can formally define the problem as follows:
\begin{definition}\textbf{Temporal Network Embedding.}
	Given a temporal network $\mathcal{G}=(\mathcal{V}, \mathcal{E}, \mathcal{T})$, temporal network embedding aims to learn a mapping function $f: \mathcal{V} \to \mathbb{R}^d$, where $d$ is the number of embedding dimensions and $d \ll |\mathcal{V}|$. The objective of the function $f$ is to model the evolution pattern of the network, including both micro- and macro-dynamics in a temporal network.
\end{definition}

\section{The Proposed Model}
\subsection{Model Overview}
Different from conventional methods which only consider the evolution of network structures, we incorporate both micro- and macro-dynamics into temporal network embedding. 
As illustrated in Figure \ref{fig:model_fig}, from a microscopic perspective (i.e., Figure \ref{fig:model_fig}(a)), we consider the establishment of edges as the chronological event and propose a temporal attention point process to capture the fine-grained structural and temporal properties for network embedding. 
The establishment of an edge (e.g., $(v_3,v_4,t_j)$) is determined by the nodes themselves and their historical neighbors (e.g., $\{v_1,v_2,\cdots\}$ and $\{v_5,v_2,\cdots\}$), where the distinct influences are captured with a hierarchical temporal attention.
From a macroscopic perspective (i.e., Figure \ref{fig:model_fig}(b)), the inherent evolution pattern of network scale constrains the network structures at a higher level, which is defined as a dynamics equation parameterized with network embedding $\mathbf{U}$ and timestamp $t$. 
Micro- and macro-dynamics evolve and derive node embeddings in a mutual manner (i.e., Figure \ref{fig:model_fig}(c)). 
The micro prediction from the historical structures indicates that node $v_3$ may link with three nodes (i.e., three new temporal edges to be established) at time $t_j$, while the macro-dynamics preserved embedding limits the number of new edges to only two, according to the evolution pattern of the network scale. 
Therefore, the network embedding learned with our proposed $\rm{M^2DNE}$ captures more precise structural and temporal properties. 

\subsection{Micro-dynamics Preserved Embedding}
With the evolution of the network, new edges are constantly being established, which can be regarded as a series of observed events. Intuitively, the occurrence of an event is not only influenced by the event participants, but also by past events. Moreover, the past events affect the current event to varying degrees. 
Thus, we propose a temporal attention point process to preserve micro-dynamics in a temporal network. 

Formally, given a temporal edge $o = (i, j, t)$ (i.e., an observed event), we parameterize the intensity of the event $\tilde \lambda_{i, j}(t)$ with network embedding $\mathbf{U} = [\mathbf{u}_i]^\top$. 
Since similar nodes  $i$ and $j$ are more likely to establish the edge $(i,j,t)$, the similarity between nodes $i$ and $j$ should be proportional to the intensity of the event that $i$ and $j$ build a link at time $t$. 
On the other hand, the similarity between the historical neighbors and the current node indicates the degree of past impact on the event $(i, j, t)$, which should decrease with time and be different from distinct neighbors. 

To this end, we define the occurrence intensity of the event $o=(i,j,t)$, consisting of the base intensity from nodes themselves and the historical influences from two-way neighbors, as follows:
\begin{align}
	\label{lambda_det}
	\tilde \lambda_{i, j}(t) & = \underbrace{g(\mathbf{u}_i,\mathbf{u}_j)}_{Base \, Intensity} \\ 
	&\nonumber 
	+ \beta_{ij}\sum_{p \in{\mathcal{H}^i(t)}}{\alpha_{pi}(t) g(\mathbf{u}_p,\mathbf{u}_j) \kappa(t-t_{p})} \\ 
	& \nonumber 
	+ \underbrace{(1-\beta_{ij})\sum_{q \in{\mathcal{H}^j(t)}}{\alpha_{qj}(t)g(\mathbf{u}_q,\mathbf{u}_i)\kappa(t-t_q)}}_{Neighbor\,Influence},
\end{align}
where $g(\cdot)$ is a function measuring the similarity of two nodes, here we define $g(\mathbf{u}_i,\mathbf{u}_j) = -||\mathbf{u}_i-\mathbf{u}_j||_2^2$, where other measurements can also be used, such as cosine similarity\footnote{Since our idea is to keep the intensity much larger when nodes $i$ and $j$ are more similar, the negative Euclidian distance is applied here as it satisfies the triangle inequality and thus can preserve the first- and second-order proximities naturally \cite{danielsson1980euclidean}.}. 
$\mathcal{H}^i(t)=\{p\}$ and $\mathcal{H}^j(t)=\{q\}$ are the historical neighbors of node $i$ and $j$ before $t$, respectively.
The term $\kappa(t-t_{p})=\exp(- \delta_{i}(t-t_{p}))$ is the time decay function with a node-dependent and learnable decay rate $\delta_i > 0$, where $t_p$ is the time of the past event $(i,p,t_p)$. 
Here $\alpha$ and $\beta$ are two attention coefficients determined by a hierarchical temporal attention mechanism, which will be introduced later. 

As the current event is stochastically excited or inhibited by past events, and the Eq. \eqref{lambda_det} may derive negative values, we apply a non-linear transfer function $f: \mathbb{R} \to \mathbb{R}_+$ (i.e., exponential function) to ensure that the intensity of an event is a positive real number.
\begin{equation}
	\lambda_{i, j}(t) = f(\tilde \lambda_{i, j}(t)).
	\label{lambda_tranfer}
\end{equation}

\subsubsection{\textbf{Hierarchical Temporal Attention}}
As mentioned before, the past events have an impact on the occurrence of the current event, and this impact may vary from past events. 
For instance, whether two researchers $i$ and $j$ collaborate on a neural network-related paper at time $t$ is usually related with their respective historical collaborators. 
Intuitively, a researcher who has collaborated with $i$ or $j$ on neural network-related papers in the past has a larger local influence on the current event $(i,j,t)$. 
Besides, if $i$'s collaborators are more experts in neural networks as a whole, his neighbors will have a larger global impact on the current event. 
Since a researcher's interest will change with the research hotspot, the influence of his neighbors is not static but dynamic. Hence, we propose a temporal hierarchical attention mechanism to capture such non-uniform and dynamic influence of historical structures. 

For the local influence from each neighbor, the term $g(p,j)= -||\mathbf{u}_{p}-\mathbf{u}_j||^2_2$ makes it likely to form an edge between nodes $i$ and $j$, if $i$'s neighbor $p$ is similar with $j$. The importance of $p$ to the event $(i,j,t)$ depends on node $i$ and changes as neighborhood structures evolve. Hence, the attention coefficient is defined as follows:
\begin{equation}
	\tilde {\alpha}_{pi}(t) = \sigma(\kappa(t-t_{p}) \mathbf{a}^\top [\mathbf{W}\mathbf{u}_{i} \oplus \mathbf{W}\mathbf{u}_{p}]),
	\label{tilde_alpha}
\end{equation}
\begin{equation}
	\alpha_{pi}(t) = \frac{\exp \left( \tilde {\alpha}_{pi}(t) \right)}{\sum_{{p}'\in \mathcal{H}^i(t)}\exp \left( \tilde {\alpha}_{p'i}(t) \right)},
	\label{alpha}
\end{equation}
where $\oplus $ is the concatenation operation. $\mathbf{a} \in \mathbb{R}^{2d}$ serves as the attention vector and $\mathbf{W}$ represents the local weight matrix. 
Here we incorporate the time decay $\kappa(t-t_p)$ so that if the timestamp $t_p$ is close to $t$, then node $p$ will have a large impact on the event $o=(i, j, t)$. 
Similarly, we can get $\alpha_{qj}(t)$ which captures the distinctive local influence from neighbors of node $j$ on the event $o$ at time $t$.

For the global impact of whole neighbors, we represent the historical neighbors as a whole with the aggregation of each neighbor information $\tilde{\mathbf{u}}_i = \sigma(\sum_{p\in\mathcal{H}^i(t)}{\alpha_{pi}(t)\mathbf{Wu}_i})$. 
Considering the global decay of influence, we average the time decay of past events with $\overline{t-t_p} = \frac{1}{|\mathcal{H}^i(t)|}\sum \limits_{p\in{\mathcal{H}^i(t)}}{(t-t_p)}$. Thus, we capture the global attention of the $i$'s whole neighbors on the current event $o$ as follows:  
\begin{equation}
	\tilde {\beta}_{i} = s(\kappa(\overline{t-t_{p}})\tilde{\mathbf{u}}_i), \quad
	\tilde {\beta}_{j} = s(\kappa(\overline{t-t_{q}})\tilde{\mathbf{u}}_j),
\end{equation}
\begin{equation}
	\beta_{ij} = \frac{\exp (\tilde {\beta}_{i} )}
	{\exp ( \tilde {\beta}_{i} ) + \exp ( \tilde {\beta}_{j})},
\end{equation}
where $s(\cdot)$ is a single-layer neural network, which takes the aggregated embedding from neighbors $\tilde{\mathbf{u}}_i$ and the average time decay of past events $\kappa(\overline{t-t_{p}}) = \exp( -\delta_{i}(\overline{t-t_{p}}))$ as input. 

Combining the two parts of attention, we can preserve the structural and temporal properties in a coupled way, as the attention itself is evolutionary with micro-dynamics in the temporal network. 

\subsubsection{\textbf{Micro Prediction}}
Until now, we define the probability of establishing an edge between nodes $i$ and $j$ at time $t$ as follows:
\begin{equation}
	p(i, j|\mathcal{H}^i(t),\mathcal{H}^j(t)) = \frac{\lambda_{i, j}(t)}{\sum\limits_{i'\in{\mathcal{H}^j(t)}}{\lambda_{i', j}(t)}+\sum\limits_{j'\in{\mathcal{H}^i(t)}}{\lambda_{i, j'}(t)}}.
	\label{prob_ij}
\end{equation}
Hence, we can minimize the following objective function to capture the micro-dynamics in a temporal network:
\begin{equation}
	\mathcal{L}_{mi} = - \sum_{t\in{\mathcal{T}}}{\sum_{(i, j, t)\in{\mathcal{E}}}{\log p(i, j|\mathcal{H}^i(t),\mathcal{H}^j(t))}}.
	\label{mi_loss}
\end{equation}

\subsection{Macro-dynamics Preserved Embedding} 
Unlike micro-dynamics driving the formation of edges, macro-dynamics describe the evolution pattern of the network scale, which usually obeys obvious distributions, i.e., the network scale can be described with a certain dynamics equation. 
Furthermore, macro-dynamics constrain the formation of the internal structure of the network at a higher level, i.e., it determines that how many edges should be generated totally by now. 
Encoding such high-level structures can largely strengthen the capability of network embeddings. 
Hence, we propose to define a dynamics equation parameterized with the network embedding, which bridges the dynamics analysis with representation learning on temporal networks.  

Given a temporal network $\mathcal{G}=(\mathcal{V}, \mathcal{E},\mathcal{T})$, we have the cumulative number of nodes $n(t)$ by time $t$. For each node $i$, it links other nodes (e.g., node $j$) with a linking rate $r(t)$ at time $t$. According to the densification power laws in network evolution \cite{leskovec2005graphs, zang2016beyond}, we have the average accessible neighbors $\zeta(n(t)-1)^\gamma$ with the linear sparsity coefficient $\zeta$ and power-law sparsity exponent $\gamma$. Hence, we define the macro-dynamics which refer to the number of new edges at time $t$ as follows:
\begin{equation}
	\Delta e'(t) = n(t)r(t)(\zeta(n(t)-1)^\gamma),
	\label{delta_e}
\end{equation}
where $n(t)$ can be obtained as the network evolves by time $t$, $\zeta$ and $\gamma$ are learnable with model optimization. 
 
As the network evolves by time $t$, $n(t)$ nodes joint in the network. At the next time, each node in the network tries to establish edges with the other $\zeta(n(t)-1)^\gamma$ nodes with a link rate $r(t)$. 

\subsubsection{\textbf{Linking Rate.}} 
Since the linking rate $r(t)$ plays a vital role in driving the evolution of network scale \cite{leskovec2005graphs}, it is dependent not only on the temporal information but also structural properties of the network. 
On the one hand, much more edges are built at the inception of the network while the growth rate decays with the densification of the network. 
Therefore, the linking rate should decay with a temporal term. 
On the other hand, the establishments of edges promote the evolution of network structures, the linking rate should be associated with the structural properties of the network. 
Hence, in order to capture such temporal and structural information in network embeddings, we parameterize the linking rate of the network with a temporal fizzling term $t^{\theta}$ and node embeddings: 
\begin{equation}
 	r(t) = \frac{\frac{1}{|\mathcal{E}|} \sum_{(i, j, t)\in{\mathcal{E}}}{\sigma(-||\mathbf{u}_i-\mathbf{u}_j||^2_2)}}{t^\theta},
 	\label{rt}
\end{equation}
where $\theta$ is the temporal fizzling exponent, $\sigma (x) = \exp (x) / (1 + \exp(x))$ is the sigmoid function. 
As the learned embeddings should well encode network structures, the numerator in Eq. \eqref{rt} models the max linking rate of the network with node embeddings, which decays over time. Hence, with Eq. \eqref{rt}, we can combine the representation learning and macro-dynamics of the temporal network. 

\subsubsection{\textbf{Macro Constraint.}}
As the network evolves, we have the sequence of the real number of edges $\mathcal{A}= \{e(t)\}_{t=t_1}^{|\mathcal{T}|}$, hence we can get the changed number of edges, denoted as $\{\Delta e(t_1)$, $\Delta e(t_2)$, $\Delta e(t_3)$, $\cdots\}$, where $\Delta e(t_i)=e(t_{i+1})-e(t_i)$. 
Then, we learn the parameters in Eq. \eqref{delta_e} via minimizing the sum of square errors:
\begin{equation}
	\mathcal{L}_{ma} = \sum_{t\in{\mathcal{T}}}{(\Delta e(t) - \Delta e'(t))^2},
\end{equation}
where $\Delta e'(t)$ is the predicted number of new edges at time $t$.

\subsection{The Unified Model}
As micro- and macro-dynamics mutually drive the evolution of the temporal network, which alternately influence the learning process of network embeddings, we have the following model to capture the formation process of topological structures and the evolutionary pattern of network scale in a unified manner: 
\begin{equation}
	\mathcal{L} = \mathcal{L}_{mi} + \epsilon \mathcal{L}_{ma},
	\label{overall_loss}
\end{equation}
where $\epsilon \in [0,1]$ is the weight of the constraint of macro-dynamics on representations learning.

\textbf{Optimization}. As the second term (i.e., $\mathcal{L}_{ma}$) is actually a non-linear least square problem, we can solve it with gradient descent \cite{marquardt1963algorithm}. However, optimizing the first term (i.e., $\mathcal{L}_{mi}$) is computationally expensive due to the calculation of Eq. \eqref{prob_ij} (i.e., $p(i, j|\mathcal{H}^i(t)$$,$ $\mathcal{H}^j(t))$). 
To address this problem, we introduce the transfer function $f$ in Eq. \eqref{lambda_tranfer} as $\exp(\cdot)$ function, thereby Eq. \eqref{prob_ij} is a Softmax unit applied to $\tilde \lambda_{i, j}(t)$, which can be optimized approximately via negative sampling \cite{mikolov2013distributed,shi2018easing}. 
Specifically, we sample an edge $(i, j, t)\in{\mathcal{E}}$ with probability proportional to its weight at each time. Then $K$ negative node pairs $(i', j, t)$ and $(i, j', t)$ are sampled. Hence, the loss function Eq. \eqref{mi_loss} can be rewritten as follows:
\begin{equation}
\fontsize{8.5pt}{8.5pt}
\begin{split}
	\mathcal{L}_{mi} & = - \sum_{t\in{\mathcal{T}}}{\sum_{(i, j, t)\in{\mathcal{E}}}} \log\sigma(\tilde \lambda_{ij}(t)) \\ & 
	- \sum_{k=1}^K{\mathbb{E}_{i'}[\log(\sigma(-\tilde\lambda_{i'j}(t)))]} - \sum_{k=1}^K{\mathbb{E}_{j'}[\log(\sigma(-\tilde\lambda_{ij'}(t)))]},
\end{split}
\end{equation}
where $\sigma (x) = \exp (x) / (1 + \exp(x))$ is the sigmoid function. Note that we fix the maximum number of historical neighbors $h$ and retain the most recent neighbors. We adopt mini-batch gradient descent with PyTorch implementation to minimize the loss function. 

The time complexity of $\rm{M^2DNE}$ is $\mathcal{O}(I\mathcal{T}(dhK|\mathcal{V}|+|\mathcal{E}|))$, where $I$ is the number of iterations and $\mathcal{T}$ means the number of timestamps. $d$ is the embedding dimension, $h$ and $K$ are the number of neighbors and negative samples. 
\section{Experiments}
In this section, we evaluate the proposed method on three datasets. Here we report experiments to answer the following questions:

\textbf{Q1. Accuracy.} Can $\rm{M^2DNE}$ accurately embed networks into a latent representation space which preserves networks structures?

\textbf{Q2. Dynamics.} Can $\rm{M^2DNE}$ effectively capture temporal information in networks for dynamic prediction tasks?

\textbf{Q3. Tendency.} Can $\rm{M^2DNE}$ well forecast the evolutionary tendency of network scale via the macro-dynamics embedding? 
\subsection{Experimental Setup}
\subsubsection{\textbf{Datasets.}}
We adopt three datasets from different domains, namely Eucore\footnote{https://snap.stanford.edu/data/}, DBLP\footnote{https://dblp.uni-trier.de} and Tmall\footnote{https://tianchi.aliyun.com/dataset/}. Eucore is generated using email data. The communications between people refer to edges and five departments are treated as labels. DBLP is a co-author network and we take ten research areas as labels. Tmall is extracted from the sales data. We take users and items as nodes, purchases as edges. The five most purchased categories are retained as labels.  
The detailed statistics of datasets are shown in Table \ref{tab:datasets}. 
\subsubsection{\textbf{Baselines.}}
We compare the performance of $\rm{M^2DNE}$ against the following seven network embedding methods. 
\begin{itemize}[leftmargin=*]
	\item \textbf{DeepWalk} \cite{perozzi2014deepwalk} performs a random walk on networks and then learns node vectors via the skip-gram model. We set the number of walks per node $w = 10$, the walk length $l = 40$. 
	\item \textbf{node2vec} \cite{grover2016node2vec} generalizes DeepWalk with biased random walks. We tune the parameters $p$ and $q$ from $\{0.25, 0.5, 1\}$.
	\item \textbf{LINE} \cite{tang2015line} considers first- and second-order proximities in networks. We employ the second-order proximity and set the number of edge samples as 100 million.
	\item \textbf{SDNE} \cite{wang2016structural} uses deep autoencoders to capture non-linear dependencies in networks. We vary $\alpha$ from $\{0.001, 0.01, 0.1\}$.
	\item \textbf{TNE} \cite{zhu2016scalable} is a dynamic network embedding model based on matrix factorization. We set $\lambda$ with a grid search from $\{0.01, 0.1, 1\}$. We take the node embeddings of last timestamp for evaluations.
	\item \textbf{DynamicTriad} \cite{zhou2018dynamic} models the triadic closure of the  network evolution. We set $\beta_0$ and $\beta_1$ with a grid search from $\{0.01, 0.1\}$, and take the node embeddings of last timestamp for  evaluations.
	\item \textbf{HTNE} \cite{zuo2018embedding} learns node representation via a Hawkes process. We set the neighbor length to be the same as ours. 
	\item \textbf{MDNE} is a variant of our proposed model, which only captures micro-dynamics in networks. (i.e., $\epsilon=0$).
\end{itemize}
\begin{table}
	\centering
	\caption{Statistics of three datasets.}
	\label{tab:datasets}
	\begin{tabular}{c|ccc}
		\toprule
		Datasets 					& Eucore				& DBLP 					& Tmall  \\
		\midrule
		\# nodes  					& 986				& 28,085				& 577,314 \\
		\# static edges 				& 24,929				& 162,451 			& 2,992,964  \\
		\# temporal edges			& 332,334			& 236,894			& 4,807,545 \\
		\# time steps				& 526				& 27					& 186 \\
		\# labels					& 5					& 10					& 5 \\
		\bottomrule
	\end{tabular}
\end{table}

\begin{table*}
	\centering
	\caption{Evaluation of network reconstruction. Pre.@K means the Precision@K.}
	\label{tab:net_reconstruction}
	\begin{tabular}{c|ccc|ccc|ccc}
		\toprule
		\multirow{2}{*}{Methods} 		& \multicolumn{3}{c|}{Eucore}  					 & \multicolumn{3}{c|}{DBLP} 						& \multicolumn{3}{c}{Tmall} \\
		\cmidrule{2-10}	
		 				& Pre.@100 		& Pre.@1000 		& AUC 			 &Pre.@100  	&Pre.@1000 		&AUC 				&Pre.@100		&Pre.@1000 		& AUC  \\
		\midrule
		DeepWalk 		&0.56 			&0.576			&0.7737 			 &0.92			&0.321			&0.9617 			&0.55			&0.455			&0.8852\\
		node2vec 		&0.73 			&0.535			&0.8157			 &0.81	 		&0.248			&0.9833  			&\textbf{0.58}  &\textbf{0.493}	&0.9755\\
		LINE			&0.58			&0.487			&0.7711			 &0.89			&0.524			&0.9859 			&0.11			&0.183			&0.8355\\
		SDNE			&0.77 			&0.747			&0.8833	 	 	 &0.88			&0.278			&0.8945 			&0.24			&0.387			&0.8934 \\
		TNE				&0.89			& 0.778 			&0.6803   		 &0.03			&0.013 			&0.9003 			&0.01  			&0.062  		&0.7278 \\
		DynamicTriad	&0.91 			&0.745 			&0.7234	   		 &0.93	  		&0.469			&0.7464				&0.27   		&0.324 			&0.9534  \\
		HTNE			&0.84			&0.776			& 0.9215 		 &0.95	  		&0.528			&0.9944	    		&0.40 			&0.404			&0.9804\\
		\midrule
		MDNE			&0.94 			&0.809 			&0.9217			 &0.97 	  		&0.543			&0.9953           	&0.25  			&0.412			&0.9853 \\
		$\rm{M^2DNE}$ 			&\textbf{0.96}	&\textbf{0.823}	&\textbf{0.9276}	 &\textbf{0.99}	&\textbf{0.553}	&\textbf{0.9964}	&0.30  			&0.431			&\textbf{0.9865}\\
		\bottomrule
	\end{tabular}
\end{table*}
\begin{table*}
	\centering
	\caption{Evaluation of node classification. Tr.Ratio means the training ratio.}
	\label{tab:node_cl}
	\begin{tabular}{c|c|c|ccccccc|cc}
		\toprule
		Datasets &Metrics &Tr.Ratio &DeepWalk &node2vec 		&LINE   		&SDNE		&TNE    		&DynamicTriad  		&HTNE   		&MDNE		&$\rm{M^2DNE}$    \\
		\midrule
		\multirow{6}{*}{Eucore} 
		& \multirow{3}{*}{Macro-F1} 
		& 40\%           &\textbf{0.1878}	&0.1575       	&0.1765  	&0.1723     &0.0954        	&0.1486        		&0.1319   		&0.1598 	&0.1365        \\
		&                           			  
		& 60\%           &0.1934        		&0.1869      	&0.1777   	&0.1834     &0.1272       	&0.1796        		&0.1731      	&0.1855		&\textbf{0.1952}        \\
		&                                         
		& 80\%           &0.2049   			&0.2022   	  	&0.1278   	&0.1987     &0.1389			&0.1979       		&0.1927 		&0.1948	&\textbf{0.2057} \\
		\cmidrule{2-12}
		& \multirow{3}{*}{Micro-F1} 
		& 40\%           &0.2089          	&0.2133         &0.2266   	&0.2129     &0.2298        	&0.2310				&0.2200   		&0.2273	&\textbf{0.2311}        \\
		&                           			 
		& 60\%           &0.2245          	&0.2400         	&0.1933   	&0.2321     &0.2377        	&0.2333       		&0.2400  		&0.2501	&\textbf{0.2533}        \\
		&                                        
		& 80\%           &0.2400		      	&0.2660    		&0.1466   	&0.2543	   	&0.2432 		&0.2400         	& 0.2672 		&0.2702	&\textbf{0.2800}   \\
		\midrule
		\multirow{6}{*}{DBLP}   
		& \multirow{3}{*}{Macro-F1} 
		& 40\%           &0.6708          	&0.6607         &0.6393  	&0.5225     &0.0580         &0.6045             &0.6768       	&0.6883	&\textbf{0.6902}        \\
		&                                          
		& 60\%           &0.6717          	&0.6681        	&0.6499   	&0.5498     &0.1429        	&0.6477             &0.6824  		&0.6915	&\textbf{0.6948}      \\
		&                                          
		& 80\%           &0.6712   			&0.6693   		&0.6513   	&0.5998     &0.1488 		&0.6642       		&0.6836 		&0.6905	&\textbf{0.6975} \\
		\cmidrule{2-12}
		& \multirow{3}{*}{Micro-F1} 
		& 40\%           &0.6653          	&0.6680         &0.6437   	&0.5517     &0.2872       	&0.6513             &0.6853   		&0.6892	&\textbf{0.6923}        \\
		&                           			  
		& 60\%           &0.6689          	&0.6737         &0.6507   	&0.5932     &0.2931        	&0.6680             &0.6857        	&0.6922	&\textbf{0.6947}       \\
		&                           			 
		 & 80\%          &0.6638   			&0.6731   		&0.6474  	&0.6423    	&0.2951 		&0.6695       		&0.6879 		&0.6924	&\textbf{0.6971} \\
		\midrule
		\multirow{6}{*}{Tmall}  
		& \multirow{3}{*}{Macro-F1} 
		& 40\%           &0.4909          	&0.5437         &0.4371  	&0.4845      &0.1069        &0.4498              &0.5481   		&0.5648	&\textbf{0.5775}        \\
		&                           			
		& 60\%           &0.4929         	&0.5455         &0.4376   	&0.4989      &0.1067        &0.4897              &0.5489    	&0.5681	&\textbf{0.5799}        \\
		&                           			
		& 80\%           &0.4953   			&0.5458   		&0.4397   	&0.5312      &0.1068 		&0.5116              &0.5493 		&0.5728	&\textbf{0.5847} \\
		\cmidrule{2-12}
		& \multirow{3}{*}{Micro-F1} 
		& 40\%           &0.5711          	&0.6041         &0.5367   	&0.5734      &0.3647        &0.5324              &0.6253   		&0.6344	&\textbf{0.6421}        \\
		&                           			
		& 60\%           &0.5734          	&0.6056         &0.5392  	&0.5788      &0.3638        &0.5688              &0.6259       	&0.6369	&\textbf{0.6438}       \\
		&                           			
		& 80\%           &0.5778   	  		&0.6066   		&0.5428  	&0.5832      &0.3642	 	&0.6072              &0.6264 		&0.6401	&\textbf{0.6465} \\
		\bottomrule
	\end{tabular}
\end{table*}
\subsubsection{\textbf{Parameter Settings}}
For a fair comparison, we set the embedding dimension $d = 128$ for all methods. The number of negative samples is set to 5. For $\rm{M^2DNE}$, we set the number of historical neighbors to 2, 5 and 2; the balance factor $\epsilon$ to 0.3, 0.4 and 0.3 for Eucore, DBLP and Tmall, respectively.

\subsection{Q1: Accuracy}
We evaluate the accuracy of the learned embeddings with two tasks, including network reconstruction and node classification. 

\subsubsection{\textbf{Network Reconstruction.}}
In this task, we train embeddings on the fully evolved network and reconstruct edges based on the proximity between nodes. 
Following \cite{chen2018pme}, for DeepWalk, node2vec, LINE and TNE, we take the inner product between node embeddings as the proximity due to their original inner product-based optimization objective. For SDNE, DynamicTriad, HTNE and our models (i.e., MDNE and $\rm{M^2DNE}$), we calculate the negative squared Euclidean distance. Then, we rank node pairs according to the proximity. As the number of possible node pairs (i.e., $|\mathcal{V}|(|\mathcal{V}|-1)/2$) is too large in DBLP and Tmall, we randomly sample about 1\% and 0.1\% node pairs for evaluation, as in \cite{ou2016asymmetric}. We report the performance in term of Precision@K and AUC.

Table \ref{tab:net_reconstruction} shows that our proposed MDNE and $\rm{M^2DNE}$ consistently outperform the baselines on AUC. On Precision@K, $\rm{M^2DNE}$ achieves the best performance on Eucore and DBLP, improving by 5.78\% and 4.73\% in term of Precision@1000, compared with the best competitors. We believe the significant improvement is because modeling macro-dynamics with $\rm{M^2DNE}$ constrains the establishment of some noise edges, so as to more accurately reconstruct the real edges in the network. Though MDNE only models the micro-dynamics, it captures fine-grained structural properties with a temporal attention point process and hence still outperforms all baselines on Eucore and DBLP. 
One potential reason that $\rm{M^2DNE}$ is not performing that well on Tmall is that the evolutionary pattern of the purchase behaviors in short term is not significant, which leads to the slightly worse performance of models for temporal network embedding (i.e., TNE, DynamicTriad, HTNE, MDNE and $\rm{M^2DNE}$). However, $\rm{M^2DNE}$ still performs better than other temporal network embedding methods in most cases, which indicates the necessity of jointly capturing micro- and macro-dynamics for temporal network embedding. 

\begin{figure}
	\centering
	\subfigure[Recall@K on Eucore]{\includegraphics[width=0.49\linewidth]{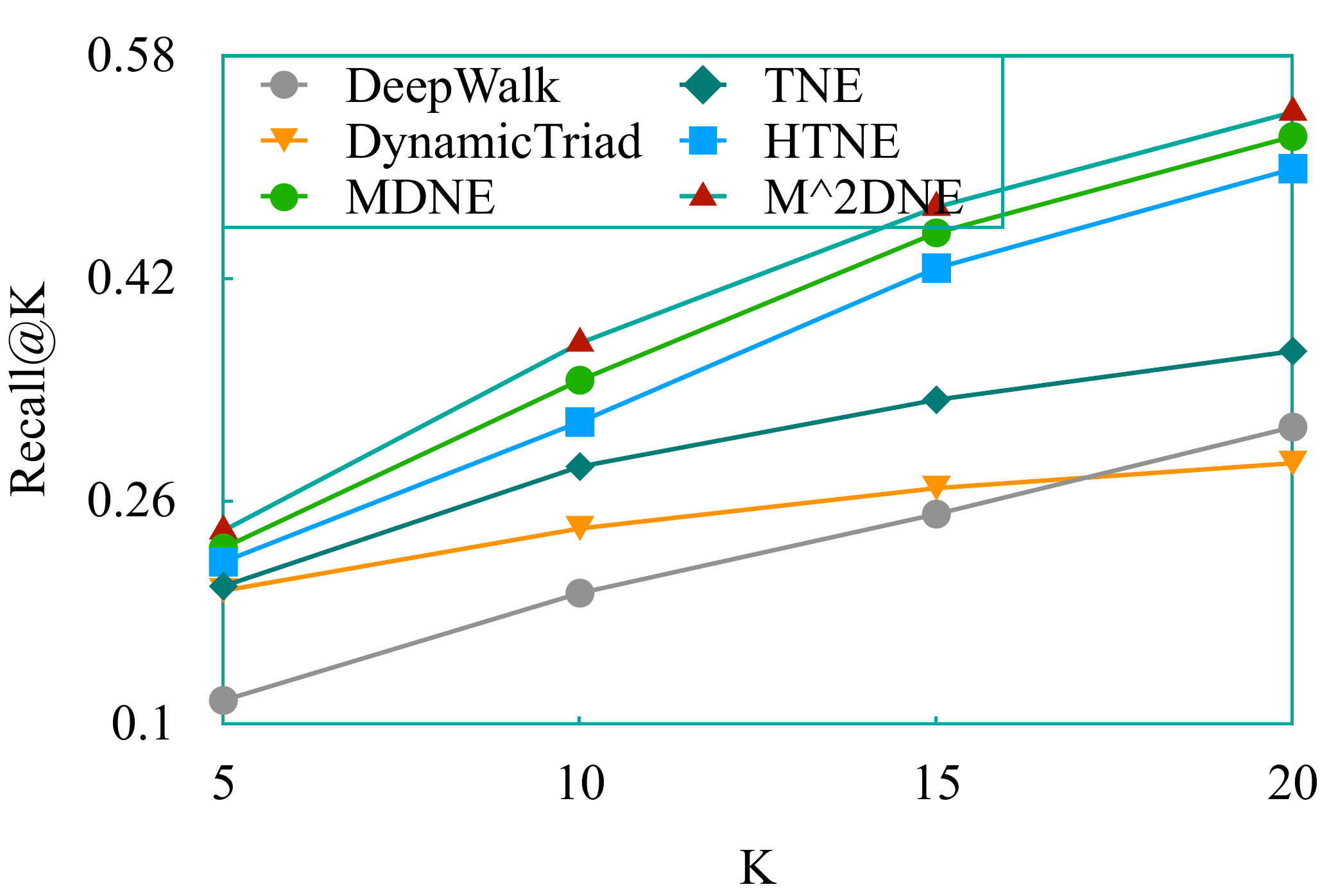}}
	\subfigure[Precision@K on Eucore]{\includegraphics[width=0.49\linewidth]{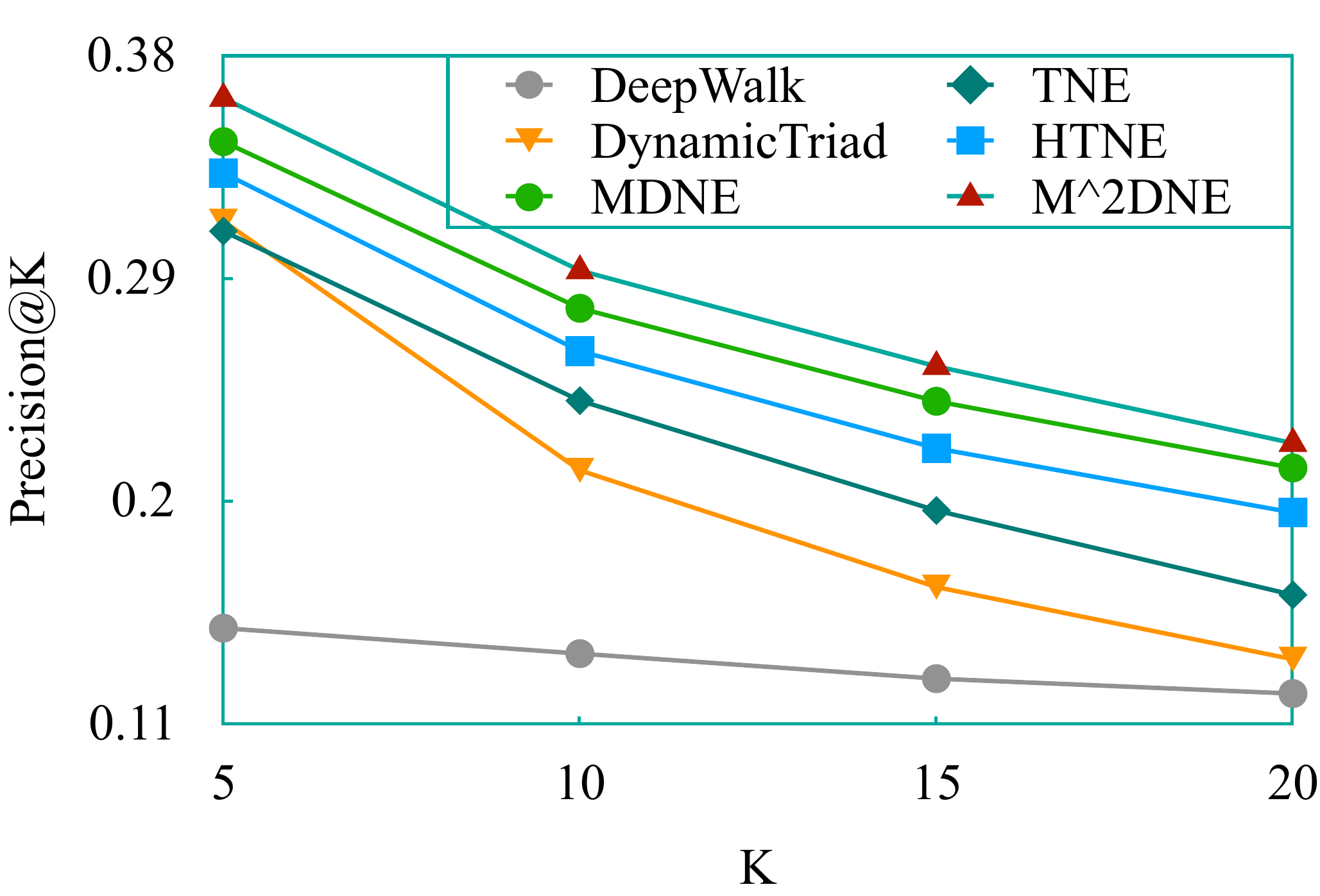}}
	\subfigure[Recall@K on DBLP]{\includegraphics[width=0.49\linewidth]{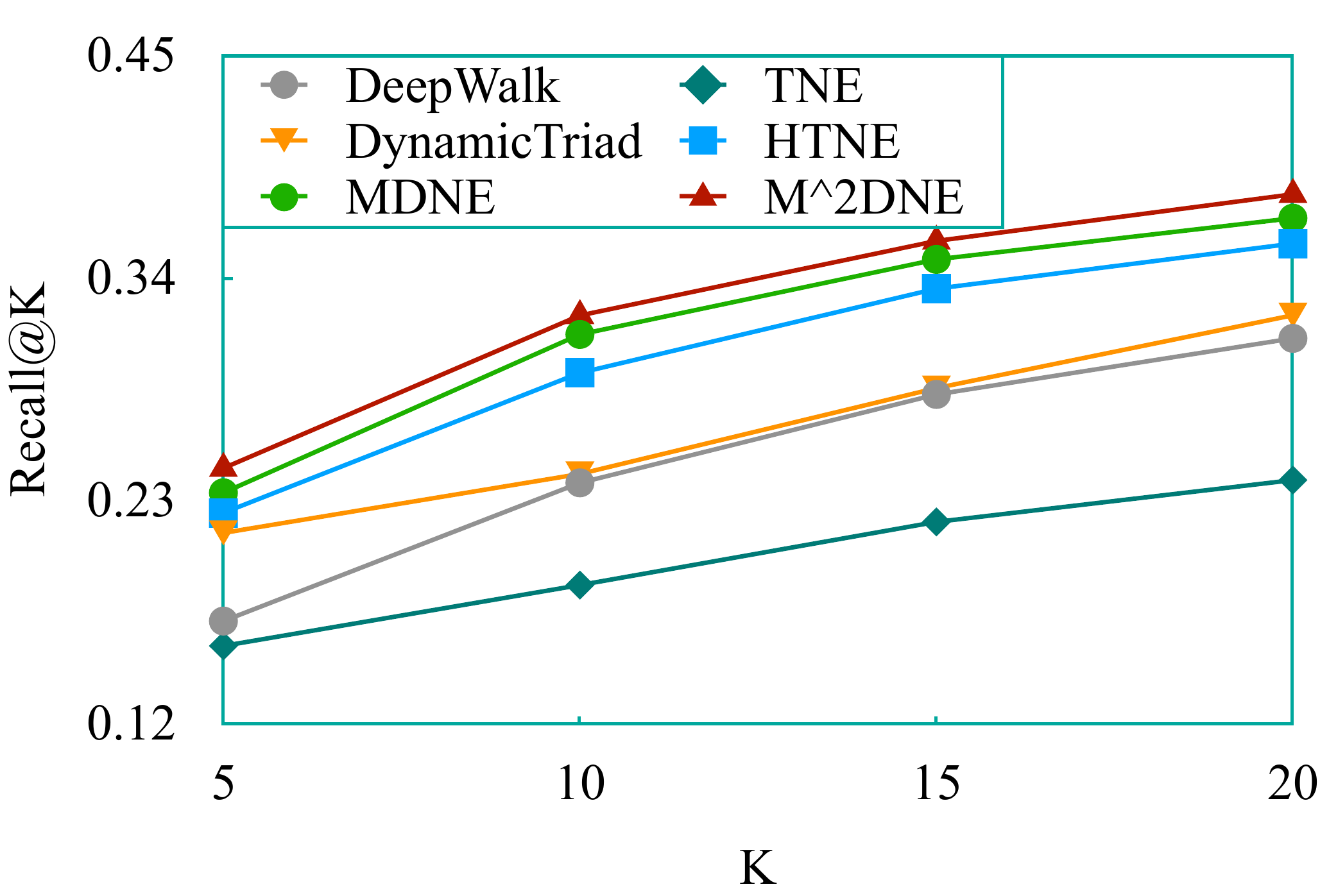}}
	\subfigure[Precision@K on DBLP]{\includegraphics[width=0.49\linewidth]{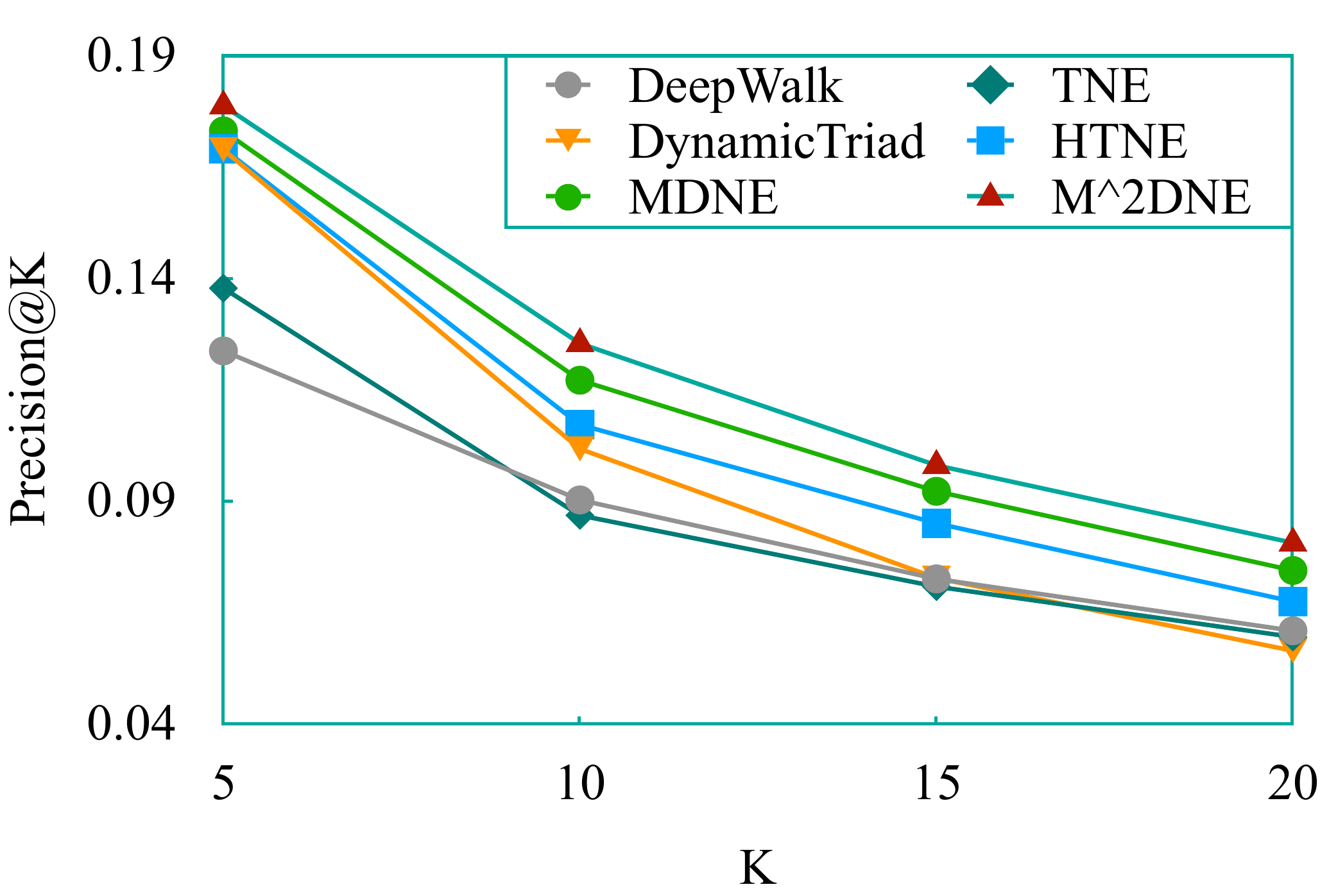}}
	\caption{Evaluation of temporal node recommendation.}
	\label{fig:tnr}
\end{figure}

\subsubsection{\textbf{Node Classification.}}
After learning the node embeddings on the fully evolved network, we train a logistic regression classifier that takes node embeddings as input features. The ratio of training set is set as 40\%, 60\%, and 80\%. We report the results in terms of Macro-F1 and Micro-F1 in Table \ref{tab:node_cl}.

As we can observe, our MDNE and $\rm{M^2DNE}$ achieve better performance than all baselines in all cases except one. Specifically, compared with methods for static networks (i.e., DeepWalk, node2vec, LINE and SDNE), the good performance of MDNE and $\rm{M^2DNE}$ suggests that the formation process of network structures preserved in our models provides effective information to make the embeddings more discriminative. In terms of methods for temporal networks (i.e., TNE, DynamicTriad and HTNE), our MDNE and $\rm{M^2DNE}$ capture the local and global structures aggregated from neighbors via a hierarchical temporal attention mechanism, which enhances the accuracy of the embeddings of structures. 
Besides, $\rm{M^2DNE}$ encodes high-level structures in the latent embedding space, which further improves the performance of classification. From a vertical comparison, MDNE and $\rm{M^2DNE}$ continue to perform best against different sizes of training data in almost all cases, which implies the stability and robustness of our models. 
\subsection{Q2: Dynamics}
We study the effectiveness of $\rm{M^2DNE}$ for capturing temporal information in networks via temporal node recommendation and link prediction. As the temporal evolution is a long term process while edges in Tmall have less significant evolution pattern with incomparable accuracy, we conduct experiments on Eucore and DBLP. 
Specifically, given a test timestamp $t$, we train the node embeddings on the network before time $t$ (not included), and evaluate the prediction performance after time $t$ (included). For Eucore, we set the first 500 timestamps as training data due to its long evolution time. For DBLP, we train the embeddings before the first 26 timestamps.

\subsubsection{\textbf{Temporal Node Recommendation.}}
For each node $v_i$ in the network before time $t$, we predict the top-$k$ possible neighbors of $v_i$ at $t$. We calculate the ranking score as the setting in network reconstruction task, and then derive the top-$k$ nodes with the highest score as candidates. 
This task is mainly used to evaluate the performance of temporal network embedding methods. However, in order to provide a more comprehensive result, we also compare our method against one popular static method, i.e., DeepWalk.

The experimental results are reported in Figure \ref{fig:tnr} with respect to Recall@K and Precision@K. We can see that our models MDNE and $\rm{M^2DNE}$ perform better than all the baselines in terms of different metrics. Compared with the best competitors (i.e., HTNE), the recommendation performance of $\rm{M^2DNE}$ improves by 10.88\% and 8.34\%in terms of Recall@10 and Precision@10 on Eucore. On DBLP, the improvement is 6.05\% and 11.69\% with respect to Recall@10 and  Precision@10. These significant improvements verify that the temporal attention point process proposed in MDNE and $\rm{M^2DNE}$ is capable of modeling fine-grained structures and dynamic pattern of the network. Additionally, the significant improvement of $\rm{M^2DNE}$ benefits from the high-level constraints of macro-dynamics on network embeddings, thus encoding the inherent evolution of the network structure, which is good for temporal prediction tasks. 
\begin{table}
	\centering
	\caption{Evaluation of temporal link prediction. }
	\label{tab:tlp}
	\begin{tabular}{c|cc|cc}
		\toprule
		\multirow{2}{*}{Methods} & \multicolumn{2}{c|}{Eucore} 	& \multicolumn{2}{c}{DBLP} 	\\
		\cmidrule{2-5}	
						&ACC. 	  			&F1					&ACC. 	  			&F1			\\
		\midrule
		DeepWalk 		&0.8444 				&0.8430				&0.7776				&0.7778 	\\
		node2vec 		&0.8591 				&0.8583				&0.8128				&0.8059 	\\
		LINE			&0.7837				&0.7762				&0.6711				&0.6756		\\
		SDNE			&0.7533	 			&0.7908				&0.6971   			&0.6867 	\\
		TNE				&0.6932	 			&0.6691	  			&0.5027   			&0.4799   	\\
		DynamicTriad	&0.6775 				&0.6611  			&0.6189   			&0.6199 	 \\
		HTNE			&0.8539 				&0.8498				&0.8123	  			&0.8157     \\
		\midrule
		MDNE			&0.8649 				&0.8585				&0.8292	  			&0.8239     \\
		$\rm{M^2DNE}$ 			&\textbf{0.8734}		&\textbf{0.8681}	&\textbf{0.8336}	&\textbf{0.8341} \\
		\bottomrule
	\end{tabular}
\end{table}
\begin{table*}
	\centering
	\caption{Evaluation of scale prediction. A.E. means the absolute error. $e'(t)$ and $e(t)$ are the predicted and real number of edges.}
	\label{tab:edge_pred}
	\begin{tabular}{c|ccc|ccc|ccc}
		\toprule
		\multirow{2}{*}{Methods} &\multicolumn{3}{c|}{Eucore} 							&\multicolumn{3}{c|}{DBLP} 						&\multicolumn{3}{c}{Tmall} \\
		\cmidrule{2-10}		
							&$e'(t)$ 	&$e(t)$		&A.E.					&$e'(t)$ 	  	&$e(t)$		&A.E.  				&$e'(t)$ 	  		&$e(t)$			&A.E. \\
		\midrule
		DeepWalk 			&444,539		&24,929		&419,610  			&335,916,746	&162,451	&335,754,295		&118,381,361,880   	&2,992,964		&118,378,368,916\\
		node2vec 			&479,583 	&24,929		&454,654 			&363,253,815	&162,451	&363,091,364   		&135,349,949,950	&2,992,964		&135,346,956,986   \\
		LINE 				&278,175		&24,929		&253,246 			&363,567,406	&162,451	&363,404,955 	  	&135,763,029,298  	&2,992,964		&135,760,036,334 \\
		SDNE 				&396,752		&24,929		&371,823				&361,748,486	&162,451	&361,586,035	  	&134,748,693,450  					&2,992,964		&134,745,700,486  \\
		TNE					&485,584  	&332,334  	&153,250			 	&389,257,712   	&236,894   	&389,020,818   		&166,630,196,186 	&4,807,545 	 	&166,625,388,641 \\
		DynamicTriad	  	&485,605   	&332,334 	&163,271			 	&394,369,570 	&236,894  	&394,132,676   	 	&165,467,872,223 	 				&4,807,545		&165,463,064,678 \\
		HTNE				&203,012		&332,334	  	&129,322				&173,501,036 	&236,894	&173,264,142       	&82,716,705,256  	&4,807,545		&82,711,897,711 \\
		\midrule
		MDNE				&203,776		&332,334	  	&128,558				&173,205,229 	&236,894	&172,968,335       	&82,702,894,887  	&4,807,545		&82,698,087,342 \\
		$\rm{M^2DNE}$ 				&349,157 	&332,334 	&\textbf{16,823}		&222,993		&236,894 	&\textbf{13,901}	&3,855,548  		&4,807,545		&\textbf{951,997}\\
		\bottomrule
	\end{tabular}
\end{table*}
\subsubsection{\textbf{Temporal Link Prediction.}} 
We learn node embedding on the network before time $t$ and predict the edges established at $t$.  
Given nodes $i$ and $j$, we define the edge's representation as $|\mathbf{u}_{i}-\mathbf{u}_{j}|$. 
We take edges built at time $t$ as positive ones, and randomly sample the same number of negative edges (i.e., two nodes share no link). And, we train a logistic regression classifier on the datasets.

As shown in Table \ref{tab:tlp}, our methods MDNE and $\rm{M^2DNE}$ consistently outperform all the baselines on both metrics (accuracy and F1). 
We believe the worse performance of TNE and DynamicTriad is due to the fact that they model the dynamics with network snapshots, which cannot reveal the temporal process of edge formation. 
Though HTNE takes the dynamic process into account, it ignores the evolutionary pattern of network size and only retains the network structures with unilateral neighborhood structures. 
Since our proposed $\rm{M^2DNE}$ captures dynamics from both microscopic and macroscopic perspectives, which embeds more temporal and structural information into node representations.  

\begin{figure}
	\centering
	\subfigure[$\frac{1}{2}|\mathcal{T}|$ on Eucore]{\includegraphics[width=0.48\linewidth]{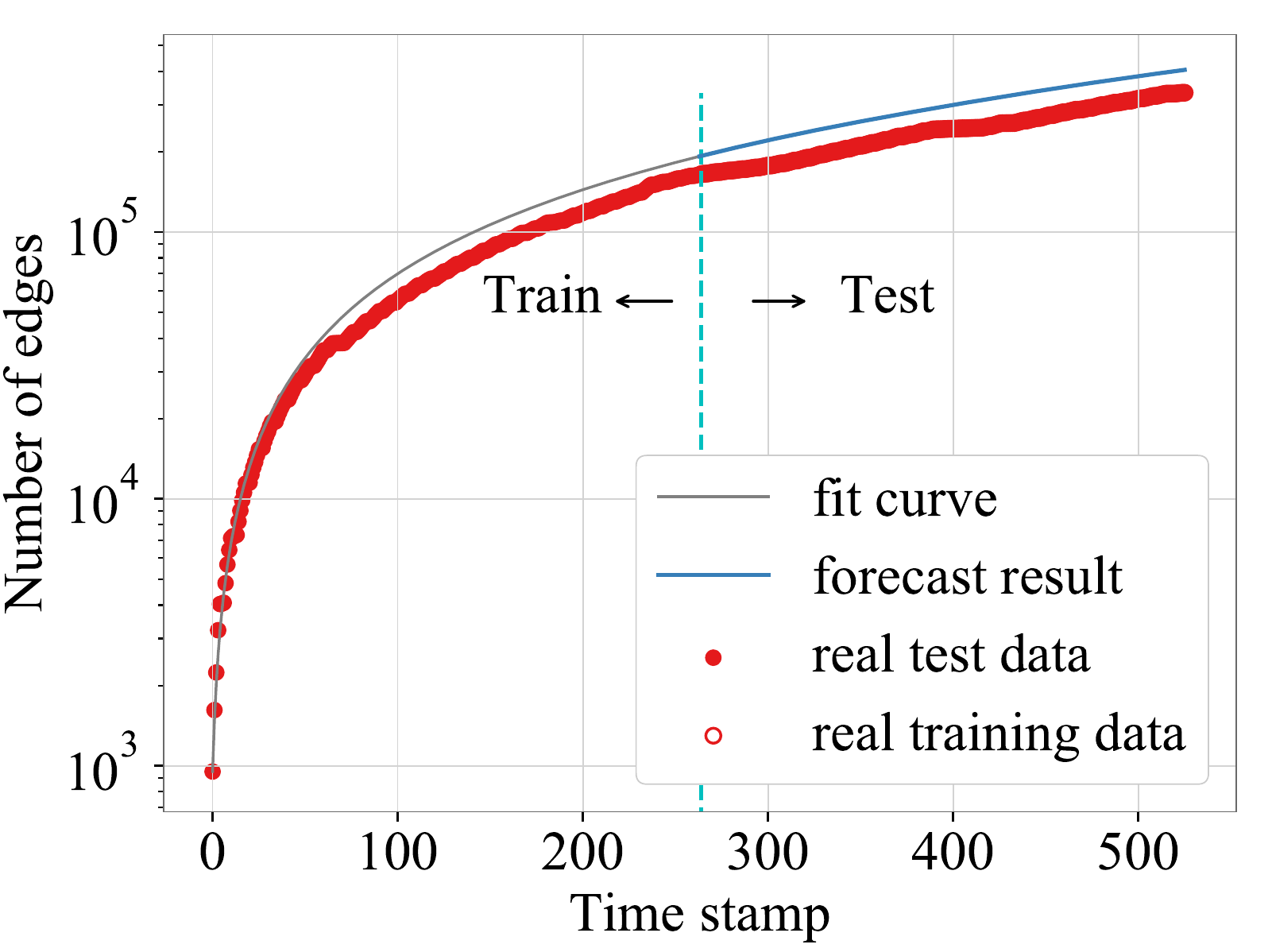}}
	\subfigure[$\frac{3}{4}|\mathcal{T}|$ on Eucore]{\includegraphics[width=0.48\linewidth]{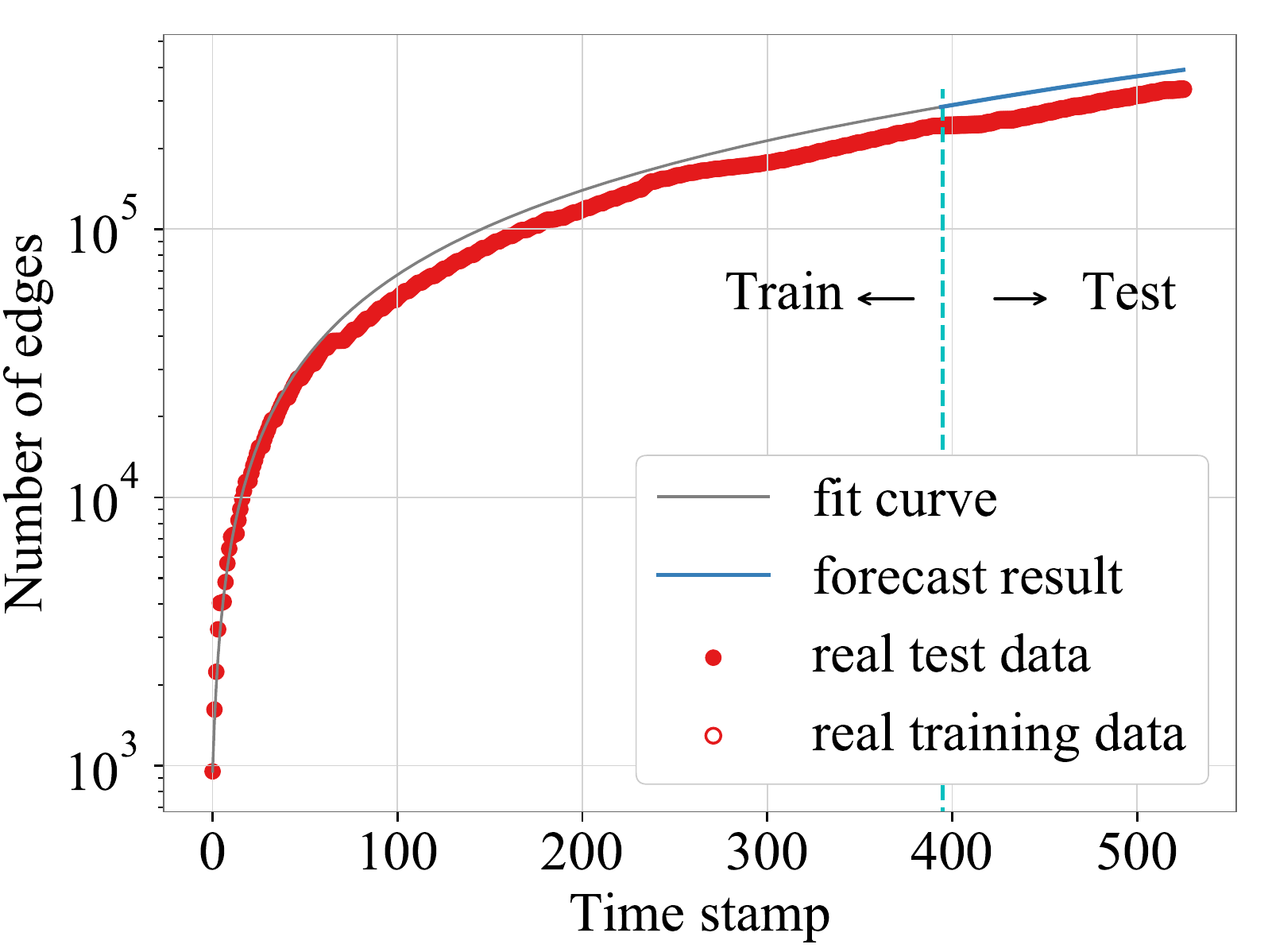}}
	\subfigure[$\frac{1}{2}|\mathcal{T}|$ on DBLP]{\includegraphics[width=0.48\linewidth]{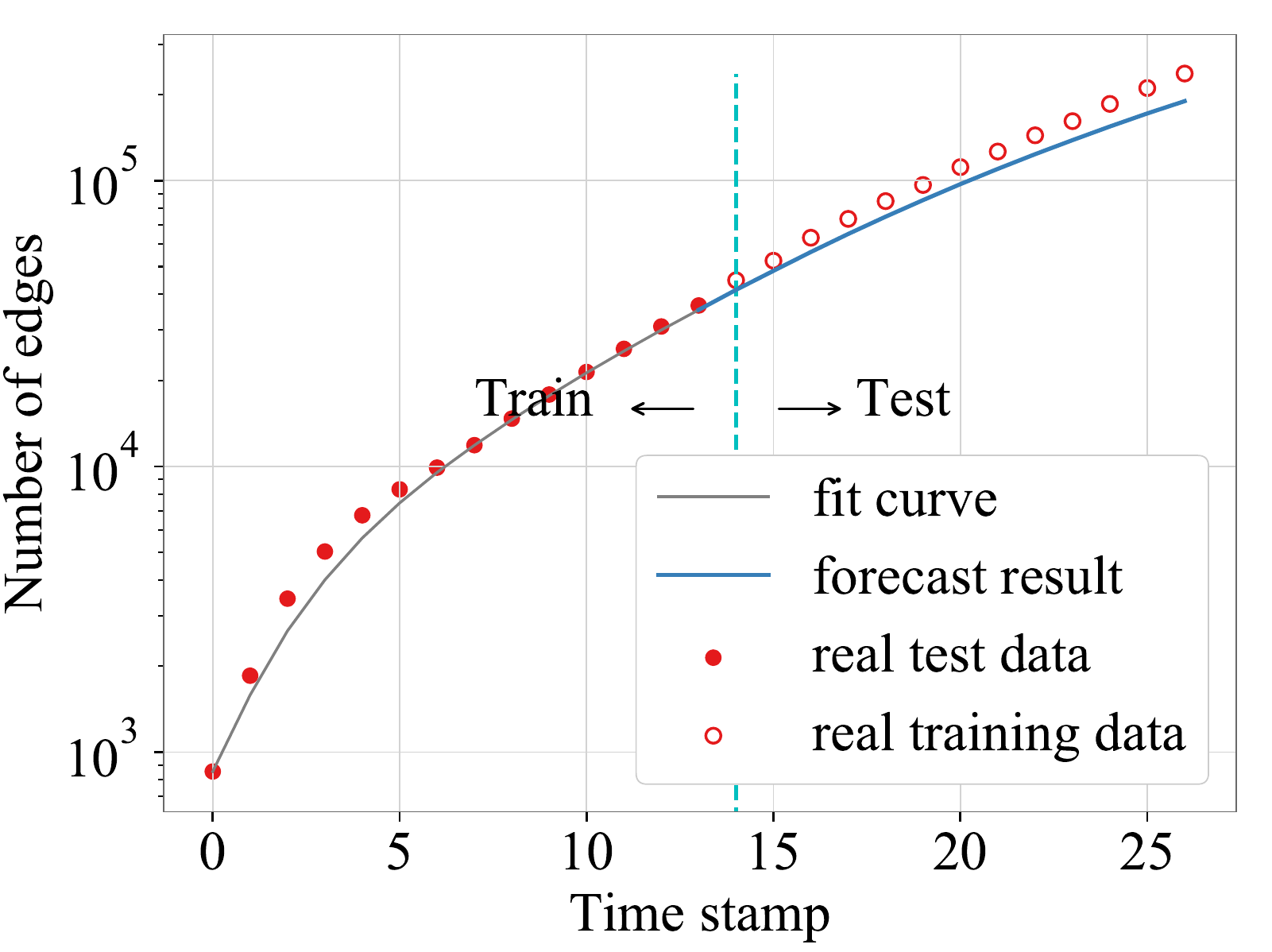}}
	\subfigure[$\frac{3}{4}|\mathcal{T}|$ on DBLP]{\includegraphics[width=0.48\linewidth]{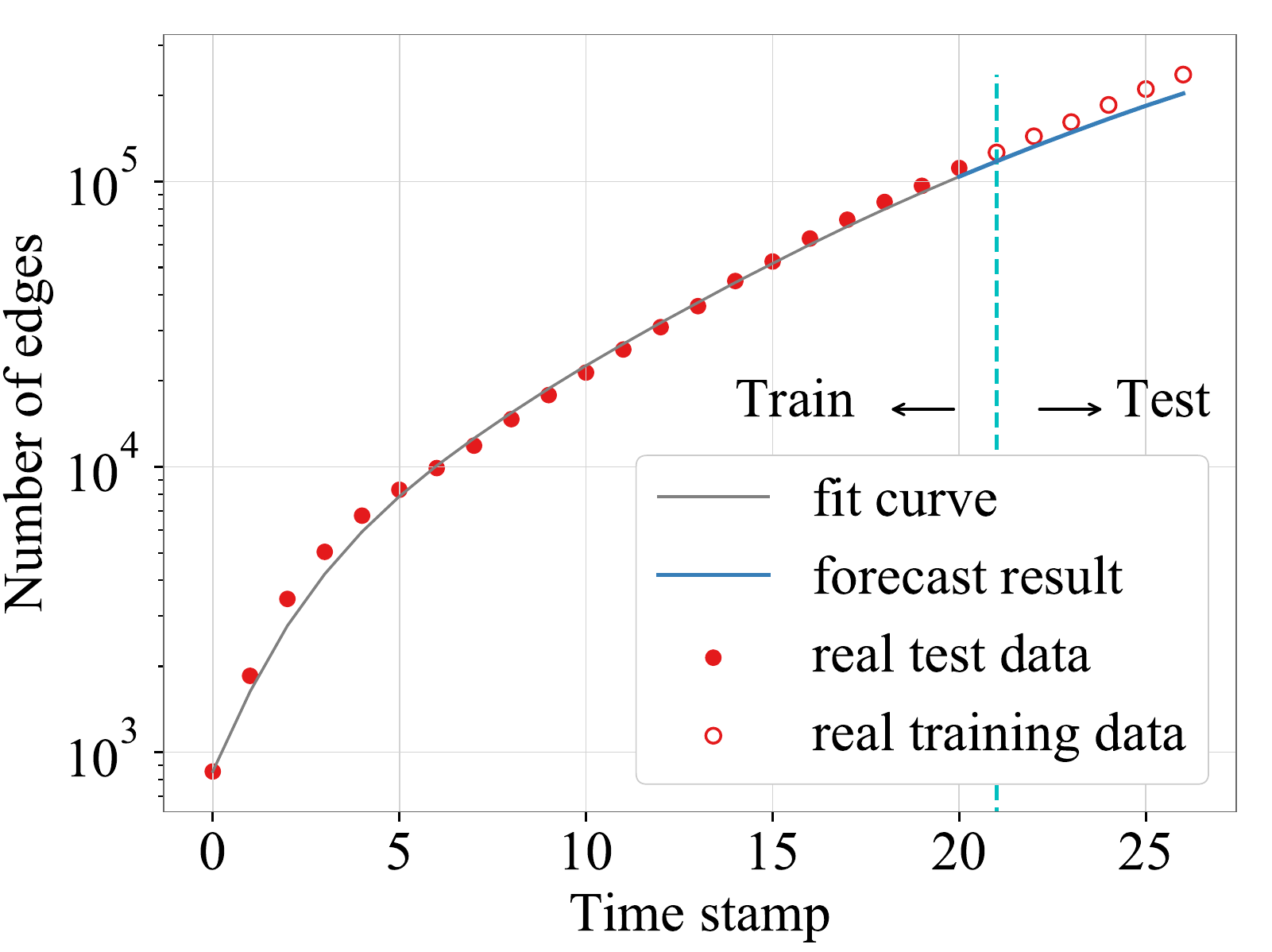}}
	\subfigure[$\frac{1}{2}|\mathcal{T}|$ on Tmall]{\includegraphics[width=0.48\linewidth]{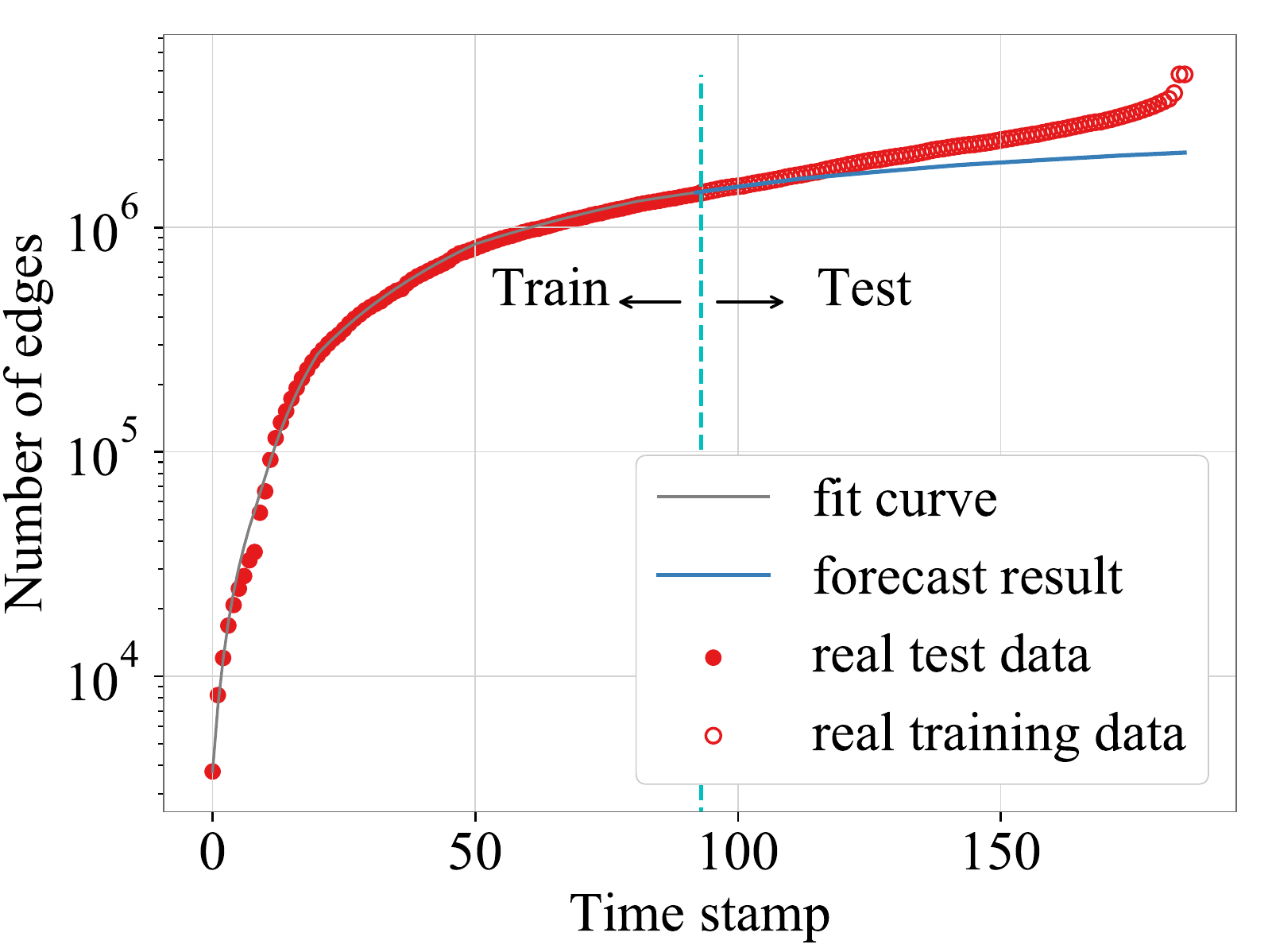}}
	\subfigure[$\frac{3}{4}|\mathcal{T}|$ on Tmall]{\includegraphics[width=0.48\linewidth]{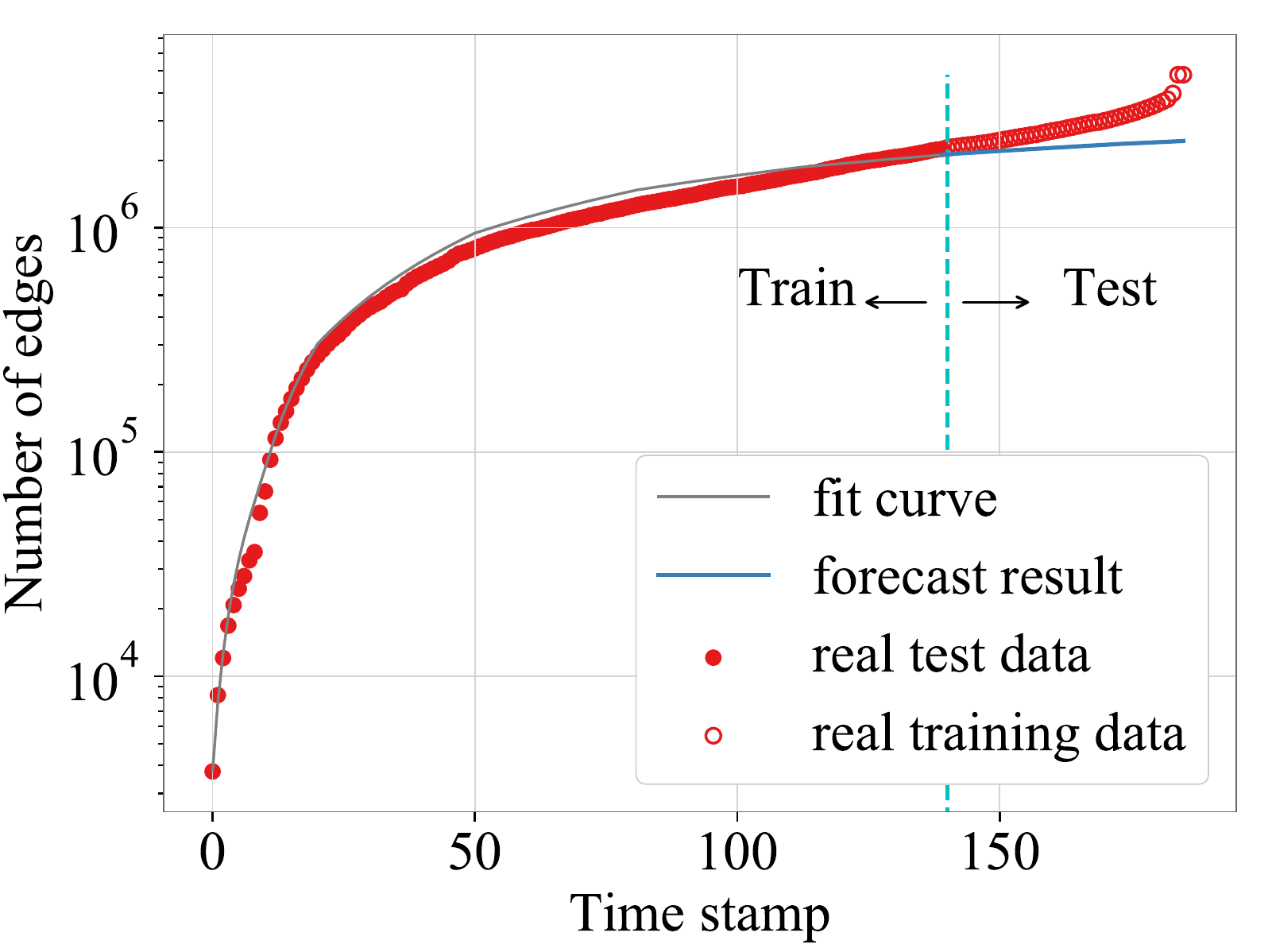}}
	\caption{Trend forecast. The red points represent real data, the filled for training, and the dashed for test. The gray lines are fit curves with training data, and the blue lines are the forecasting results of the number of edges.}
	\label{fig:size_fore}
\end{figure}
\subsection{Q3: Tendency}
As $\rm{M^2DNE}$ captures both the micro- and macro-dynamics, our model can not only be applied to some traditional applications, but also some macro dynamics analysis related applications. Here, we perform scale prediction and trend forecast tasks.

\subsubsection{\textbf{Scale Prediction}}
In this task, we predict the number of edges at a certain time. 
We train node embeddings on networks before the first 500, 26 and 180 timestamps on Eucore, DBLP and Tmall, respectively. Then we predict the cumulative number of edges by next time. Since none of the baselines can predict the number of edges, we calculate a score for each node pairs, defined as $\phi(i, j) = \sigma(\mathbf{u}_i \mathbf{u}_j)$. If $\phi(i, j) > 0.5$, we count an edge connected nodes $i$ and $j$ at next time. While in our method, we can directly calculate the number of edges with Eq. \eqref{delta_e}.

The edge prediction results with respect to absolute error (i.e., A.E.) are reported in Table \ref{tab:edge_pred}. $e(t)$ is the real number of edges and $e'(t)$ is the prediction. Note that we only count static edges for models designed for static network embedding, while for models designed for dynamic networks, we count temporal edges, that is why we have different $e(t)$ in Table \ref{tab:edge_pred}. 
Apparently, $\rm{M^2DNE}$ can predict the number of edges much more accurately, which verify that the embedding parameterized dynamics equation (i.e., Eq. \eqref{delta_e}) is capable of modeling the evolution pattern of the network scale.  
The prediction errors of the comparison methods are very large, because they can not or can only capture the local evolution pattern of networks, while ignore the global evolution trend.

\subsubsection{\textbf{Trend Forecast}}
Different from the scale prediction at a certain time, we forecast the overall trend of the network scale in this task. 
Given a timestamp $t$, we learn node embeddings (i.e., $\mathbf{U}$) and parameters (i.e., $\theta, \zeta$ and $\gamma$) on the network before time $t$ (not included). Based on the learned embeddings and parameters, we forecast the evolution trend of network scale and draw the evolution curves. 
Since none of the baselines can forecast the network scale, we only perform this task with our method w.r.t. different training timestamp. 
We set the training timestamp as $\frac{1}{2}|\mathcal{T}|$ and $\frac{3}{4}|\mathcal{T}|$ ($\mathcal{T}$ is the timestamp sequence) and forecast the  remaining with Eq. \eqref{delta_e}. 

As shown in Figure \ref{fig:size_fore}, $\rm{M^2DNE}$ can well fit the number of edges, with network embeddings and parameters learned via the dynamics equation (i.e., Eq. \eqref{delta_e}), which proves that the designed linking rate (i.e., Eq. \eqref{rt}) well bridges the dynamics of network evolution and embeddings of network structures.  
Moreover, as the training timestamp increases (i.e, from $\frac{1}{2}|\mathcal{T}|$ to $\frac{3}{4}|\mathcal{T}|$), the prediction errors decrease, which is consistent with common sense, i.e., more training data will help the learned node embeddings better capture the evolutionary pattern.  
We also notice that the forecast accuracy is slightly worse on Tmall, since the less significant evolutionary pattern of purchase behaviors in short term.

\begin{figure}
	\centering
	\subfigure[\# historical neighbors]{\includegraphics[width=0.485\linewidth]{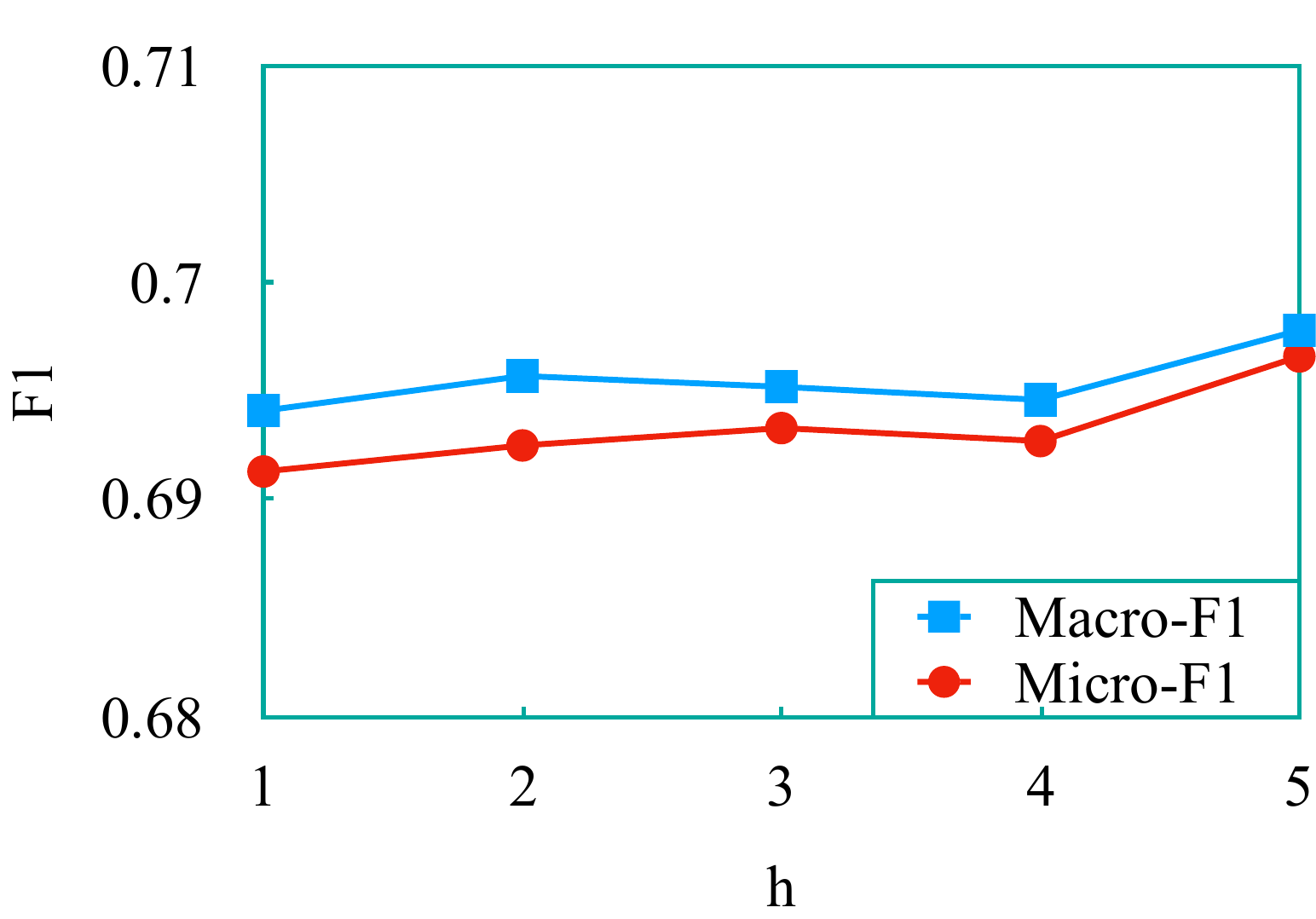}}
	\subfigure[\# negative samples]{\includegraphics[width=0.485\linewidth]{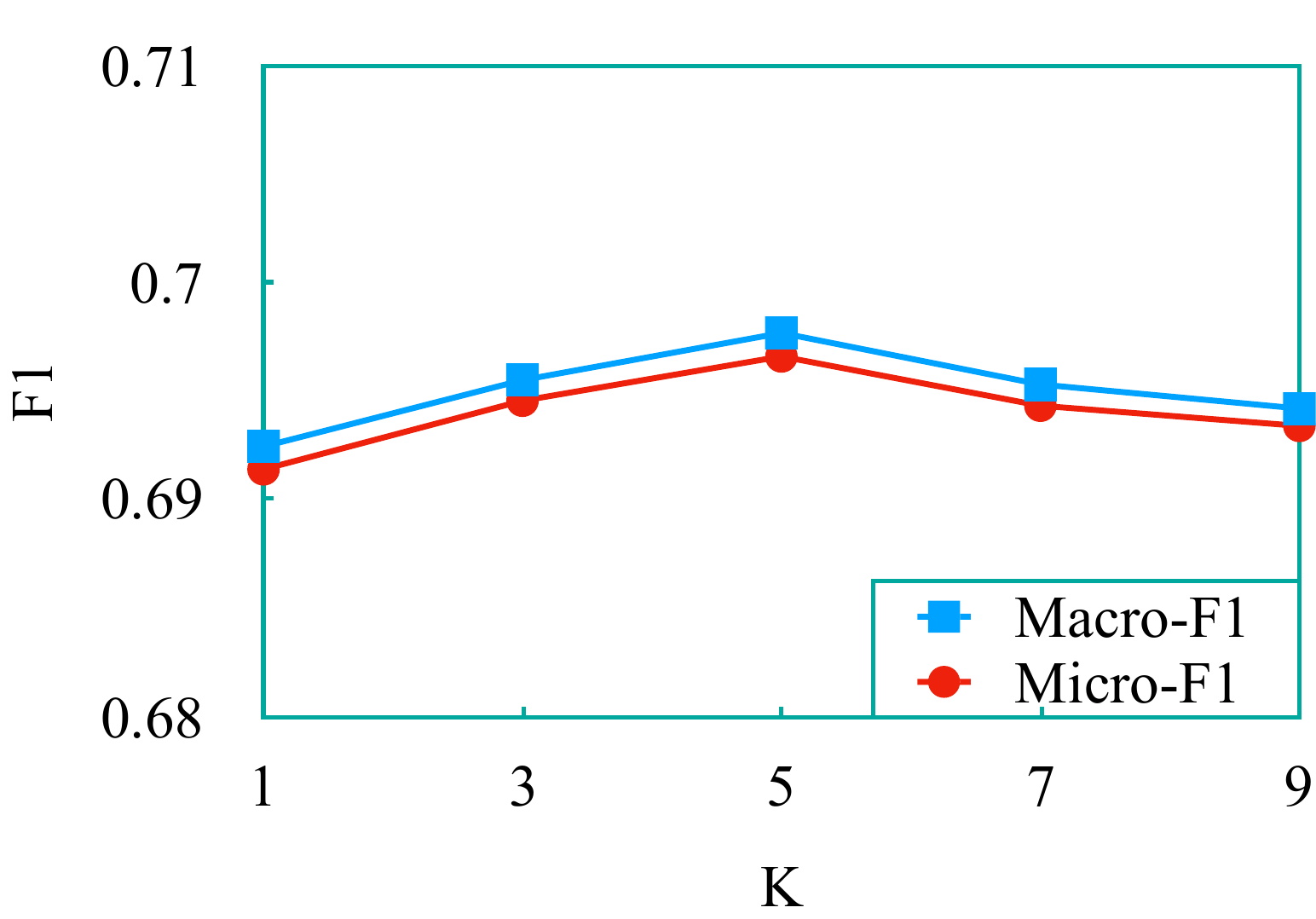}}
	\caption{Parameter analysis on DBLP.}
	\label{fig:para_ana}
\end{figure}
\subsection{Parameter Analysis}
Since the number of historical neighbors $h$ determines the captured neighborhood structures and the number of negative samples $K$ affects model optimization, we explore the sensitivity of these two parameters on DBLP dataset. 
\subsubsection{\textbf{Number of Historical Neighbors}}
From Figure \ref{fig:para_ana}(b), we notice that $\rm{M^2DNE}$ performs stably and achieves the highest F1 score at 5. Since historical neighbors models the formation of network structures, the number of historical neighbors influences the performance of model. To strike a balance between performance and complexity, we set the number of neighbors as a small value.
\subsubsection{\textbf{Number of Negative Samples.}} 
As shown in Figure \ref{fig:para_ana}(b), the performance of $\rm{M^2DNE}$ improves with the increase in the number of negative samples, and then achieves the best performance once the dimension of the representation reaches around 5. Overall, the performance of $\rm{M^2DNE}$ is stable, which proves that our model is less sensitive to the number of negative samples.

\section{Conclusion}
In this paper, we make the first attempt to explore temporal network embedding from microscopic and macroscopic perspectives. We propose a novel temporal network embedding with micro- and macro-dynamics ($\rm{M^2DNE}$), where a temporal attention point process is designed to capture structural and temporal properties at a fine-grained level, and a general dynamics equation parameterized with network embedding is presented to encode high-level structures by constraining the inherent evolution pattern. 
Experimental results demonstrate that $\rm{M^2DNE}$ outperforms state-of-the-art baselines in various tasks. 
One future direction is to generalize our model to incorporate the shrinking dynamics of temporal networks. 

\section{Acknowledgments}
This work is supported in part by the National Natural Science Foundation of China (No. 61772082, 61702296, 61806020), the National Key Research and Development Program of China (2017YFB0803304), the Beijing Municipal Natural Science Foundation (4182043), and the 2018 and 2019 CCF-Tencent Open Research Fund. This work is supported in part by NSF under grants III-1526499, III-1763325, III-1909323, SaTC-1930941, and CNS-1626432. This work is supported in part by the NSF under grants CNS-1618629, CNS-1814825, CNS-1845138, OAC-1839909 and III-1908215, the NIJ 2018-75-CX-0032.

\bibliographystyle{ACM-Reference-Format}
\bibliography{MMDNE_bibliography}

\end{document}